# One Model, Two Precisions: Does Disagreement Between Quantized Copies Tell You When a Small Model Is Wrong?

*Cross-precision disagreement as an uncertainty signal for selective prediction on local multiple-choice question answering, with matched-noise, option-order and ensemble controls*

Mohammed Nayeem Khan
mohammed.nayeem.khan@intel.com
mnknayeem@gmail.com

Allan Kevin T
allan.kevin.t@intel.com

Intel Corporation

**Abstract**

Local language models often exist at several precisions, and quantization is known to change individual answers. This study asks whether the disagreement between two precisions of the *same* weights, an 8-bit copy that answers and a lower-precision copy that reads the same prompt, works as a per-question error score for abstention, escalation and re-asking. The cost is one extra forward pass over fewer bytes, with no sampling and no training. We simulate round-to-nearest quantization with per-group scales (3 to 6 bits, groups of 32 to 256) on Qwen3-0.6B and Qwen3-1.7B and score 8,398 multiple-choice questions from ARC-Easy, ARC-Challenge, MMLU and HellaSwag subsets. Error detection is measured by AUROC, risk-coverage curves and ECE on a held-out half, with paired bootstrap intervals over questions. For Qwen3-0.6B, the entropy of the 8-bit answer reaches AUROC 0.682; the best disagreement score, selected on a development half, reaches 0.684 (+0.001 [-0.009, +0.011]), and combining it with the confidence features adds +0.009 [+0.002, +0.015], mostly on ARC. On Qwen3-1.7B the selected score is -0.001 [-0.009, +0.006] from entropy and the combination +0.003 [-0.001, +0.006] on the test half. Gaussian weight noise at matched error energy is about as informative as rounding (0.6B only), so we find no evidence that the signal is specific to quantization, and shuffling the features within task and entropy cells removes the gain. One extra 8-bit pass that shifts the option order does better (+0.049 [+0.031, +0.067] over entropy) at about a third more weight traffic; routing to a larger model on free entropy beats both extra-pass scores at equal total traffic. After logistic recalibration the 0.6B confidences are calibrated within a few points of ECE. A deployment recipe and the cases where the method does not help are given.

**Key findings (every number recomputed from the cached option logits)**

1. On Qwen3-0.6B, the entropy of the 8-bit answer detects wrong answers with AUROC 0.682 [0.667, 0.698] on the held-out half; the dev-selected precision-disagreement score gets 0.684 [0.666, 0.700], difference +0.001 [-0.009, +0.011] (p=0.760).
2. On Qwen3-1.7B (n = 4,199 test questions) the dev-selected precision-disagreement score (w6g128, pdrop) differs from entropy by -0.001 [-0.009, +0.006]. The logistic combination with the confidence features differs by +0.003 [-0.001, +0.006]. The 1.7B model was run with 3 low-precision copies against 9 for the 0.6B model. With all questions and five-fold cross-fitting the combination gains +0.004 [+0.001, +0.006] over the confidence features.
3. Adding precision disagreement to the three confidence features raises the Qwen3-0.6B AUROC by +0.009 [+0.002, +0.015]. On this model the gain is positive on ARC-Easy and ARC-Challenge (+0.020 [+0.011, +0.027], +0.024 [+0.012, +0.035]) and inside the noise on MMLU and HellaSwag. For Qwen3-1.7B the pattern differs: ARC-Easy +0.000 [-0.003, +0.003], HellaSwag +0.018 [+0.007, +0.028].
4. Matched-noise control (0.6B only, 4-bit energy): 1 - p_low[answer] has AUROC 0.550 for rounding and 0.597 on average over 3 noise seeds (range 0.531 to 0.649) for per-tensor Gaussian weight noise. The seed spread is as wide as the gap, so no evidence is found that the signal is specific to quantization.
5. The signal is best for the copy closest to the original among those run, and no copy between 6.25 and 8.25 effective bits was tried. It degrades with damage and reverses for 3-bit copies (JSD AUROC 0.410), because a near-random copy disagrees most with confident answers.
6. Nearly cost-matched (1.00 against 0.76 units): one extra 8-bit pass with shifted options gives AUROC 0.732 [0.716, 0.748] (+0.049 [+0.031, +0.067] over entropy); the logistic combination with confidence gives 0.742. A shifted-option pass on a low-precision copy was not run.
7. After a logistic recalibration on a development half, the 0.6B ECE falls from 0.181 (raw max-probability) to 0.032 (entropy) and 0.020 (with precision disagreement); these intervals overlap. For 1.7B the entropy-only map leaves 0.073, against 0.021 for the three confidence features.
8. Routing 0.6B to 1.7B: option order routes best at equal escalated share, but at equal total weight traffic (signal pass included) the free entropy ranking is ahead of both extra-pass scores.

## 1 Introduction

A person who runs a small language model on their own machine usually has no second opinion about a given answer. The model returns a letter, and the only free evidence of how much to trust it is the probability it assigned. That probability is informative but crude, and the better-known alternatives cost more: sampling several reasoning chains and voting [23], asking the model to grade itself [21, 25], or training a verifier. Local users, however, often hold the same model at more than one precision, or can make a lower-precision copy cheaply. Quantization is applied to nearly every local deployment, it changes some answers [3, 4, 2], and it is reported to change them most where the full model was already unsure [1].

This suggests a cheap experiment: run the question through the deployed copy, run it again through a lower-precision copy of the same weights, and read the disagreement between the two as an uncertainty score. The extra cost is one forward pass over fewer bytes, with no sampling and no training. The paper asks whether that disagreement predicts when the deployed copy is wrong, whether it adds anything to the entropy and confidence the model already provides, and what it should be compared with at equal cost.

The answers come from simulated quantization (round-to-nearest with per-group scales, 3 to 6 bits, group sizes 32 to 256) of Qwen3-0.6B and Qwen3-1.7B, on 8,398 multiple-choice questions from ARC-Easy, ARC-Challenge, a subset of MMLU and a subset of HellaSwag. Each claim is scoped to that setting. The controls are Gaussian weight noise at matched error energy, other bit-widths and group sizes, and an option-order disagreement that costs one extra 8-bit pass. Further checks use ensembles of copies and a domain-shift test over MMLU subject categories.

The main results hold for small models, simulated quantization and these four benchmarks. For the 0.6B model, the entropy of the 8-bit answer detects wrong answers with AUROC 0.682. The best low-precision disagreement score, chosen on a development half, gets 0.684 on the held-out half, +0.001 [-0.009, +0.011] from entropy. Adding it to the confidence features gives +0.009 [+0.002, +0.015], a small gain that on this model is concentrated in the ARC tasks. For the 1.7B model the same score is -0.001 [-0.009, +0.006] from entropy and the combination +0.003 [-0.001, +0.006]. Gaussian weight noise of the same energy is about as informative as quantization on the 0.6B model, so we found no evidence that the signal is specific to rounding. Option-order disagreement costs one extra 8-bit pass, about a third more weight traffic than the low-precision pass (0.76 against 1.00 units). It gives +0.049 [+0.031, +0.067] and is the stronger signal in these experiments, though a shifted-option pass on a low-precision copy, at equal cost, was not run. After a logistic recalibration the 0.6B confidences are calibrated to within a few points, so a recalibrated confidence is the baseline to beat.

Quantization changes confidence and flips individual answers [1, 3, 4, 2]. Ensembles, weight perturbation and cheap ensembles are standard uncertainty tools [8, 9, 14, 15], selective prediction, calibration and cascades are established [18, 11, 27], and option-order inconsistency is a known weakness of language models on multiple choice [31, 32]. The paper claims none of that. The searches we ran (logged with the paper) found no per-question use of the disagreement between two precisions of one model as an error score for selective prediction, with controls that separate it from generic weight noise and from option-order disagreement at similar cost. This paper adds that use, the controls and the measured outcome: a small gain that is not specific to quantization, and a stronger signal from option-order disagreement at about a third more weight traffic. The search covered a paper index and abstracts only, so a missed precedent is possible.

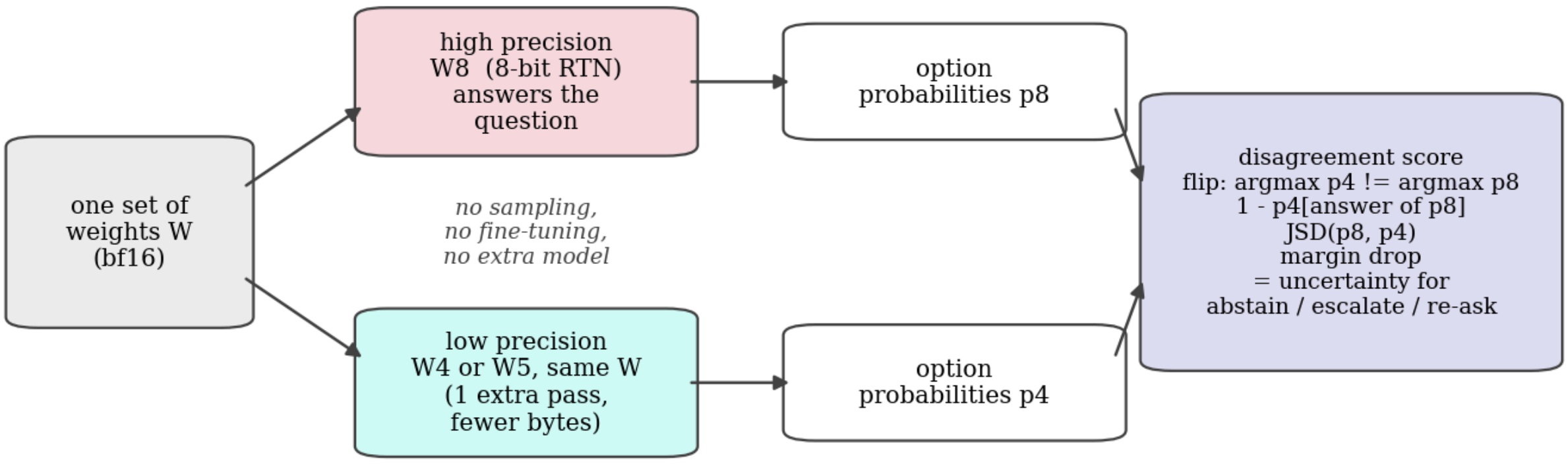

**Figure 1.** The method. The same weights are used twice. The 8-bit copy answers the question; a second, lower-precision copy of the same weights reads the same prompt, and the way its option probabilities differ from the 8-bit copy's becomes an uncertainty score. The method needs no sampling, extra model or training.

## 2 Related work

### 2.1 What quantization does to confidence and to individual answers

Proskurina et al. [1] study the confidence of GPTQ-quantized models. They find that 4-bit quantization lowers the confidence assigned to the true label, with effects that vary across model families and scales, and they report that the loss falls disproportionately on samples where the full model was already low-confidence. That last observation is the premise of this paper: if quantization moves the items the model is unsure about, the movement itself may carry information. Dutta et al. [3] show that equal accuracy can hide many answers that flip from right to wrong and back, and propose KL divergence and flip rates as distances between a compressed model and its parent. A later study [4] relates the argmax flip rate to total variation across 802 compressed or perturbed copies of 19 models. Rababah et al. [2] measure "correctness agreement" between a base model and its quantized variants from 8 to 2 bits and find behavioural divergence at moderate quantization even when accuracy looks preserved. Studies of trustworthiness and reasoning under compression [34, 35, 36] document further changes, and one of them [36] reports that token positions with high KL divergence between quantized and full-precision outputs coincide with high next-token entropy. That anticipates the correlation with entropy found below. Across this work the flip or divergence is a *property of the compressed model*, averaged over a benchmark, and used to judge a quantizer. Here the same quantity is read *per question* and used to decide about one answer.

### 2.2 Ensembles, weight perturbation and cheap ensembles

Deep ensembles [8] and Monte Carlo dropout [9] are the standard ways to get a model-disagreement uncertainty; both pay for several forward passes, and ensembles also pay for several trained models. Cheaper variants share most parameters: BatchEnsemble [14], Packed-Ensembles [15], singular value ensembles for foundation models [16], and training-free Bayesianization of LoRA adapters through a weight variance [17]. Abe et al. [13] caution that the uncertainty and robustness benefits of deep ensembles are not always beyond what a single larger network provides, and Ovadia et al. [12] show how uncertainty quality changes under dataset shift. The approach studied here belongs to the same family, with members produced by rounding the weights of one trained model to fewer bits. Nothing is trained and nothing is stored beyond the quantized copy a local user may already hold.

### 2.3 Uncertainty and selective prediction for language models

The maximum softmax probability is the usual baseline [10], and modern networks are known to be miscalibrated [11]. For language models, Kadavath et al. [21] show that larger models are well calibrated on multiple-choice questions when the format is right. Semantic entropy [22], self-consistency [23], an information-theoretic epistemic score [24] and self-evaluation for selective prediction [25] need sampled generations or extra prompts. A large study [7] compares token-probability, numerical and linguistic verbal uncertainty on MMLU-Pro across 80 models from 0.6B parameters, including quantized variants. Two recent papers examine ambiguous questions. One finds that estimators based on the predictive distribution or on an ensemble of models degrade to close-to-random performance on them [5]. The other finds that uncertainty metrics predict errors better on unambiguous instances [6]. This is a reason to prefer benchmarks with a single intended answer, as used here, when comparing signals. Selective classification, the reject option, goes back to Geifman and El-Yaniv [18, 19]; Traub et al. [20] list requirements for multi-threshold evaluation, which motivate the use of AURC and coverage-wise accuracy next to AUROC. Abstention in language models is surveyed in [26]. A recent taxonomy [41] separates input-level, parameter-level, token-level and decoding-level uncertainty in language models and groups estimators into Bayesian, ensemble, consensus and single-pass families. In that vocabulary, perturbing the weights of one model, as done here, is a parameter-level, ensemble-style estimator.

### 2.4 Option order, routing and multiple precisions

Multiple-choice answers depend on option position and order [31, 32]. Permuting the options and comparing answers is therefore a natural disagreement signal and is the closest alternative at similar cost used here. The same idea of marginalising out the order of items in a prompt is used for list ranking [42]. Routing and cascades send easy queries to a small model and hard ones to a larger one [27, 28, 29, 30]. This paper studies the uncertainty score that drives the router. Any-precision LLMs [33] store several bit-widths of one model in roughly the memory of the largest, which makes a second precision cheap to hold; this paper asks what a second precision can say about the first one's answers.

## 3 Method

### 3.1 Models, items and scoring

The models are Qwen3-0.6B and Qwen3-1.7B [37], bf16 checkpoints from the Hugging Face hub (Apache-2.0 per the model metadata). The decoder blocks of the two models hold 440 M and 1409 M weights in the seven projections that the quantizer touches. The four benchmarks are ARC-Easy and ARC-Challenge [38] (2,376 and 1,172 questions: the full test splits after dropping questions with fewer than two or more than five options), a fixed random subset of MMLU [39] (2,850 questions, 50 from each of 57 subjects) and a fixed random subset of HellaSwag [40] (2,000 items), which makes 8,398 questions per model. Every question is shown as a plain-text prompt with lettered options, zero-shot, ending in `Answer:`. The answer is the option letter with the largest next-token logit among the letters that exist for that question, and the option probabilities p are the softmax over those logits. No chat template and no sampling are involved, so a forward pass gives everything needed.

The aim is an uncertainty score per question, so the label that matters is whether the answer given by the deployed (high-precision) copy is wrong, and everything below is about predicting that label.

### 3.2 The simulated quantizer

The quantization is simulated in memory. The study uses no packed kernels, GGUF files or GPU integer kernels, and it claims no wall-clock saving. For each of the q, k, v, o, gate, up and down projections of every decoder layer, the input dimension of a weight row is cut into groups of $g$ consecutive weights. In each group the code is `q = clip(round((w - min) / s), 0, 2^b - 1)` with scale `s = (max - min) / (2^b - 1)`, and the weight is replaced by `q s + min` (rounded to bf16 when it is written back). This is plain round-to-nearest with a per-group scale and offset, asymmetric, with no clipping search, no calibration data and no error compensation. Embeddings, the output head and the normalisation weights stay in bf16. The storage cost of such a format is $b + 32/g$ bits per weight if scale and offset are fp16, which we call the effective bits (for example 4.25 for 4-bit, group 128, and 5 for 4-bit, group 32). Dequantised values are stored in bf16, which is itself a rounding of about 8 bits of mantissa. The 8-bit copy therefore differs a little from the original checkpoint, and that difference is measured below because it is the noise floor of the whole approach.

We use the 8-bit, group-128 copy as the high-precision member, the model that answers the question. The low-precision member is the same weights at 3, 4, 5 or 6 bits and group sizes 32 to 256. 9 copies for the 0.6B model and 3 for the 1.7B model were run (Appendix B). No 7-bit copy was run, so no copy lies between 6.25 effective bits and the 8-bit reference. The 16-bit original is also run, so that the 8-bit copy can be compared with it.

### 3.3 Signals

Write p8 for the option probabilities of the high-precision copy, $y$ = argmax p8 for its answer, and pL for the probabilities of a low-precision copy on the same prompt. Scores are oriented so that a larger value means more likely wrong, which was fixed in advance, and an AUROC below 0.5 is reported as it is.

- **Confidence baselines** (free, from p8 only): entropy of p8; one minus the maximum probability; negative log-ratio of the top two probabilities (the margin) [10].
- **Precision disagreement** (one extra pass at low precision): *flip* = [argmax pL differs from y]; *pdrop* = 1 - pL[y], the low-precision copy's probability that the high-precision answer is wrong; *JSD*, the Jensen-Shannon divergence between p8 and pL (natural log, at most 0.693); *KL* from p8 to pL; *margin drop* = margin(p8) - margin(pL).
- **Option-order disagreement** (one extra pass at high precision): the options are shifted cyclically by one position, the logits are mapped back to the original options, and the same five quantities are computed against that second distribution [31, 32]. This costs one extra 8-bit pass (1.00 units against 0.76 for the 6-bit copy), so it is the closest comparison at similar cost that was run; a shifted-option pass on a low-precision copy was not run. Three shifts together give a four-way vote.
- **Ensembles of copies**: the fraction of four 4-bit copies (group sizes 32, 64, 128, 256) whose answer differs from y, and the same for three noisy copies (below).

### 3.4 Controls

The matched-noise control asks whether quantization noise is just one kind of weight perturbation: if it is, any perturbation of the same size would do the same job. For each weight tensor, Gaussian noise with the same root-mean-square error as the 4-bit group-128 quantizer is added to the 16-bit weights, either with one variance per tensor or with one variance per group of 128 (so that the signal-to-noise ratio matches group by group). Noise seeds are fixed and several are run. The matched noise has the same energy but no dependence on the weight values, so if it is as informative as quantization, the signal is generic and we should say so. Only the 4-bit energy was run for the noise copies (0.6B model).

The other controls are the bit-width sweep (3 to 6 bits), the group-size sweep (32 to 256), the two model sizes and the four tasks. MMLU categories serve as domains for a shift test, and the 8-bit versus 16-bit comparison gives a noise floor.

### 3.5 Selection, splits and statistics

The questions are split at random into a development half and a test half (seed 14). The low-precision configuration and the single disagreement score are chosen on the development half by AUROC among 9 configurations and five scores for the 0.6B model; the 1.7B model has only 3 copies, so its choice is among 3 configurations. In both models the chosen copy is the one with the most bits that was run, so the choice sits at the edge of the grid. Everything headline is reported on the test half (n = 4,199), where the choice cannot flatter itself. Fitted combinations are logistic regressions of “answer is wrong” on standardised features, fitted on the development half and scored on the test half. Analyses that need all questions (per task, per subject, strata, shift, routing) use five-fold cross-fitting with whole folds held out. There the low-precision configuration comes from the development-half choice, which leaks a little (one choice among 9 configurations). Uncertainty is a percentile bootstrap over questions (200 to 400 resamples depending on the table), and differences between two scores on the same questions use the paired bootstrap, which resamples the same questions for both. The p-value in a paired comparison is the two-sided share of resamples on the wrong side of zero, floored at 1/(B+1). It is not corrected for the number of comparisons in a table.

### 3.6 Metrics

AUROC treats wrong answers as the positive class, and 0.5 is chance. Partial AUROC restricts to the false-alarm rates up to 0.2 (correct answers flagged), rescaled so that 0.5 is chance, which is the region that matters when abstention is expensive. For risk-coverage, the questions are sorted from most to least certain, the top share (the coverage) is answered, and the error rate on the answered ones is the risk. AURC is the mean of that risk over all coverages, and accuracy at 50% and 80% coverage is read off the same curve [18, 20]. ECE uses 15 equal-width bins after the confidence has been recalibrated by a logistic map fitted on the development half, and the raw max-probability ECE is reported next to it, with the Brier score alongside. Cost is counted in weight bytes read per question, in units of one 8-bit group-128 pass (8.25 bits per weight); a pass at $b$ bits and group $g$ costs $(b + 32/g)/8.25$ of that. Wall-clock time on a shared GPU would mean little and is not used; the quantization is simulated anyway.

## 4 How far apart are the precisions?

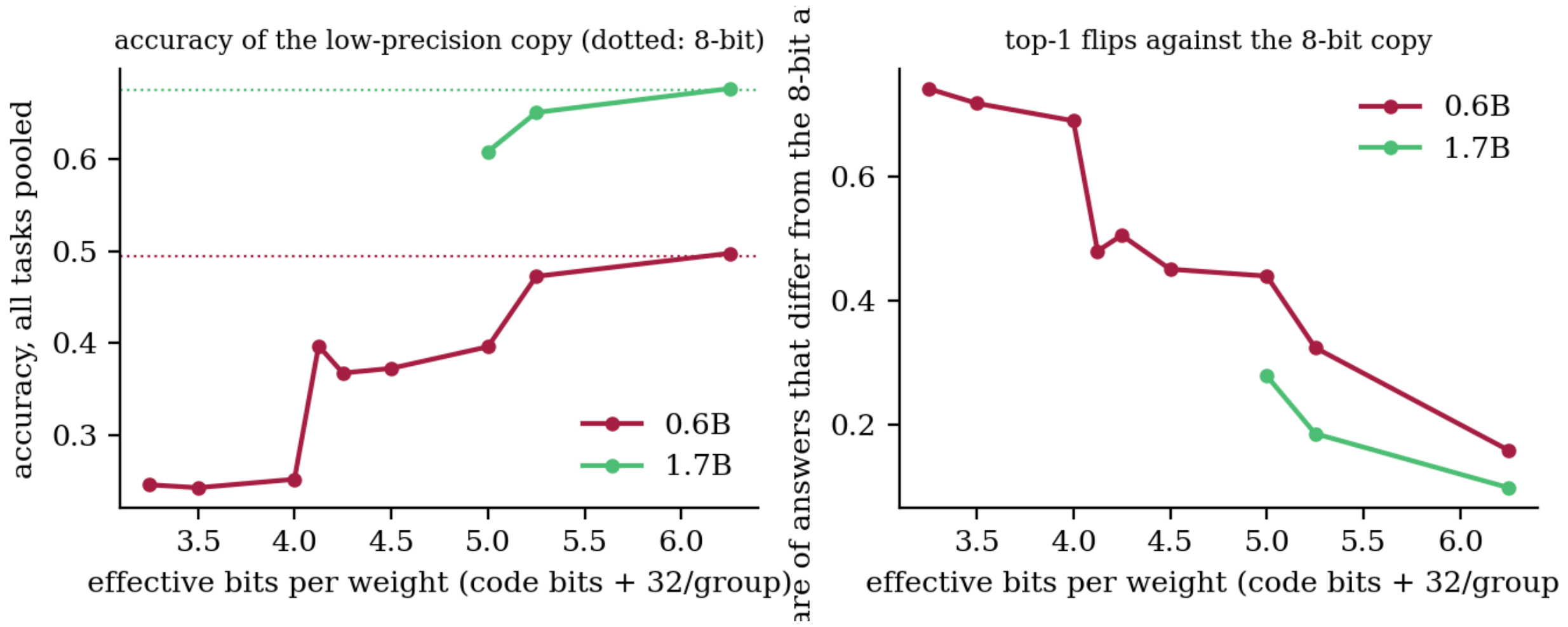

**Figure 2.** Accuracy (left) and answer flips against the 8-bit copy (right) for every low-precision copy, against effective bits per weight. Points of one curve are the bit-width and group-size combinations that were run; dotted lines are the 8-bit accuracy.

Figure 2 and Table 1 show how far the copies drift as bits are removed. For the 0.6B model the 8-bit copy changes 5.0% of the answers relative to the 16-bit checkpoint, and its accuracy is 49.5% against 49.6% for the original. That 5.0% is the noise floor of everything below, because the 8-bit copy is the reference: part of any measured disagreement is bf16 rounding. Below 8 bits the damage grows quickly for a model of this size. At 6.25 effective bits (w6g128) the copy keeps 49.8% accuracy and flips 15.8% of the answers. At 4 bits with group 128 the accuracy is 36.8% and 50.6% of the answers differ from the 8-bit copy's. At 3 bits the copy is close to random on all four tasks (24.6% overall, where chance on these four-way tasks is about 25%). Finer groups reduce the damage at a price in bits: 4 bits at group 32 (5 effective bits) flips 44.0%.

For a disagreement signal this sets up a tension. A copy that is almost identical to the original flips only a few answers, and those flips are partly rounding noise. A copy that has lost a quarter of its accuracy flips so many answers that the flip no longer says anything about the question. The useful range, if there is one, lies between.

**Table 1.** Accuracy of each copy by task (original option order) and the share of questions on which the copy's answer differs from the 8-bit copy's. g is the group size; effective bits include the 32 bits of scale and offset per group.

| model | copy | eff. bits | ARC-E | ARC-C | MMLU | HellaSwag | all | flips vs 8-bit |
|---|---|---|---|---|---|---|---|---|
| Qwen3-0.6B | 16-bit original | 16 | 67.1% | 49.7% | 41.9% | 39.5% | 49.6% | - |
| Qwen3-0.6B | w8g128 | 8.25 | 66.8% | 49.8% | 41.8% | 39.6% | 49.5% | - |
| Qwen3-0.6B | w6g128 | 6.25 | 67.2% | 51.2% | 42.4% | 38.8% | 49.8% | 15.8% |
| Qwen3-0.6B | w5g128 | 5.25 | 63.4% | 49.8% | 38.9% | 38.4% | 47.3% | 32.4% |
| Qwen3-0.6B | w4g32 | 5.00 | 51.3% | 38.1% | 32.9% | 36.1% | 39.6% | 44.0% |
| Qwen3-0.6B | w4g128 | 4.25 | 44.5% | 37.6% | 32.7% | 32.9% | 36.8% | 50.6% |
| Qwen3-0.6B | w3g128 | 3.25 | 25.6% | 23.0% | 22.9% | 26.9% | 24.6% | 74.3% |
| Qwen3-1.7B | 16-bit original | 16 | 86.8% | 72.4% | 54.9% | 60.7% | 67.7% | - |
| Qwen3-1.7B | w8g128 | 8.25 | 86.6% | 71.7% | 54.6% | 60.7% | 67.5% | - |
| Qwen3-1.7B | w6g128 | 6.25 | 86.2% | 72.2% | 54.4% | 61.9% | 67.6% | 9.8% |
| Qwen3-1.7B | w5g128 | 5.25 | 84.3% | 68.6% | 50.9% | 60.2% | 65.1% | 18.5% |
| Qwen3-1.7B | w4g32 | 5.00 | 80.5% | 64.8% | 48.3% | 52.8% | 60.8% | 27.8% |

## 5 Does disagreement predict errors?

Table 2 and Figure 3 give the first result. On the test half of the 0.6B model, the entropy of the 8-bit answer detects wrong answers with AUROC 0.682 [0.667, 0.698]. The disagreement score chosen on the development half, which is `pdrop` against the w6g128 copy, gets 0.684 [0.666, 0.700]; the paired difference from entropy is +0.001 [-0.009, +0.011] (p=0.760). On this model and these tasks, one extra low-precision pass produces a score about as informative as the free entropy, and not measurably more. One minus the maximum probability gives 0.680, the margin 0.671.

The table has two points that need care. First, the choice of disagreement score changes the result. The flip (a yes/no event) and the JSD (a divergence over all options) are weaker than 1 - p_low[answer], the probability that the low-precision copy gives to the high-precision answer. The last score is graded, and it behaves like a second confidence reading of the same answer; the flip throws that grading away, and the JSD also counts disagreement about options nobody picked. Second, when the low-precision copy is poor the sign of the relation can reverse: at 3 bits the JSD has an AUROC below 0.5 (0.410), because a near-random copy disagrees most with a sharp, confident answer. The rank correlation between the entropy of p8 and the JSD is -0.411 at 3 bits and 0.405 at 6 bits, so a disagreement score measures a mixture of the model's own uncertainty and the damage done by the quantizer, with the mix set by how rough the quantizer is.

The option-order score is the closest comparison at similar cost that was run: one extra pass at 8-bit precision, which costs 1.00 units against 0.76 for the low-precision pass. Its graded form, 1 - p'[answer], reaches 0.732 [0.716, 0.748], which is +0.049 [+0.031, +0.067] above entropy (p=0.003). Its flip and JSD forms are below entropy. Combining the three confidence features with the option-order features by logistic regression gives 0.742 [0.727, 0.758]; with the precision features instead it gives 0.691 [0.674, 0.707], +0.009 [+0.002, +0.015] from entropy (p=0.020). The pooled picture is therefore a small gain from precision disagreement and a larger one from option-order disagreement, on this model.

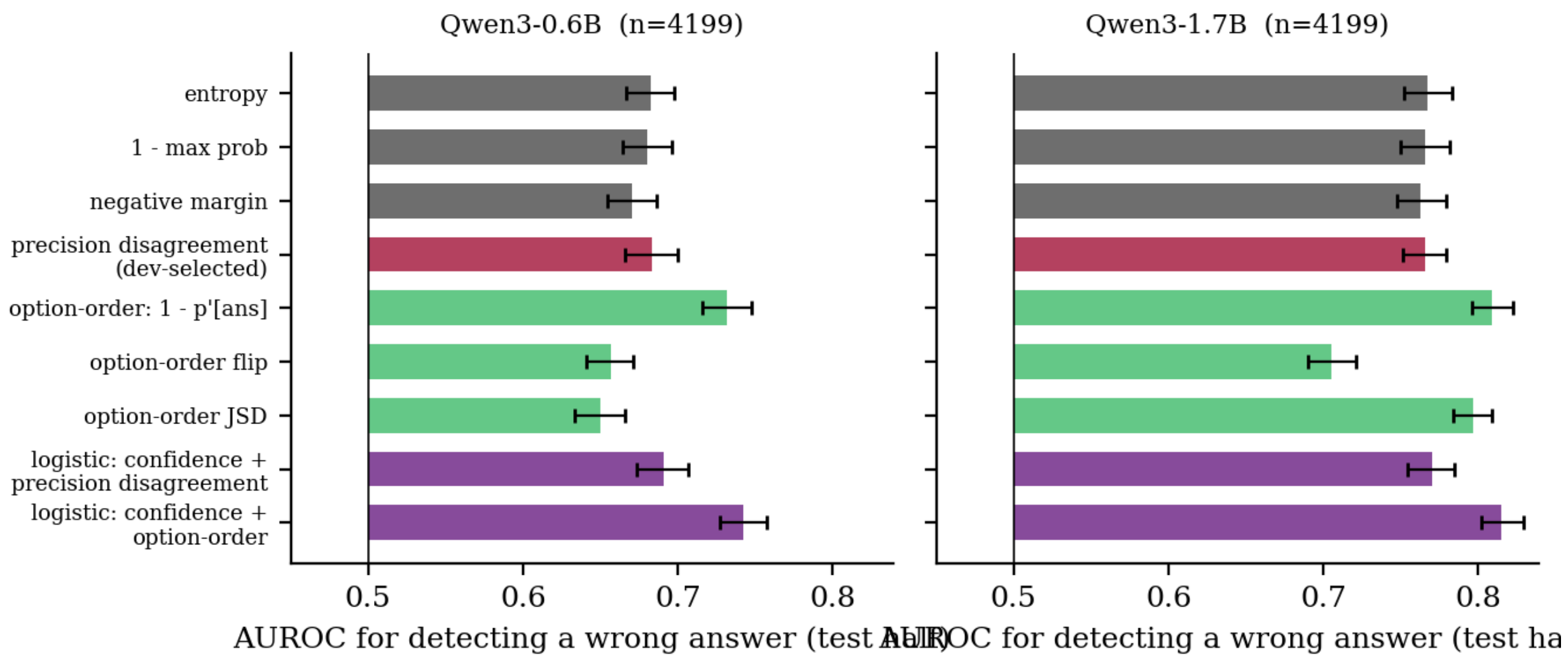

**Figure 3.** AUROC for detecting a wrong answer on the test half, with 95% bootstrap intervals over questions. From the top: free confidence scores of the 8-bit copy; the disagreement score with the best AUROC on the development half (one extra low-precision pass); option-order disagreement (one extra 8-bit pass); logistic combinations fitted on the development half.

**Table 2.** Error detection on the test half (n = 4,199 per model). Extra cost is in units of one 8-bit pass. The difference column is the AUROC minus the entropy AUROC on the same questions with a paired bootstrap interval. pAUC is the partial AUROC at false-alarm rates up to 0.2 (McClish-standardised, 0.5 is chance). AURC: lower is better.

| model | score | extra cost | AUROC [95% CI] | difference from entropy | p | pAUC | AURC |
|---|---|---|---|---|---|---|---|
| Qwen3-0.6B | entropy of 8-bit answer | 0.00 | 0.682 [0.667, 0.698] | - | - | 0.556 | 0.341 |

| model | score | extra cost | AUROC [95% CI] | difference from entropy | p | pAUC | AURC |
|---|---|---|---|---|---|---|---|
| Qwen3-0.6B | 1 - max probability | 0.00 | 0.680 [0.664, 0.696] | -0.002 [-0.007, +0.003] | p=0.367 | 0.555 | 0.342 |
| Qwen3-0.6B | negative margin | 0.00 | 0.671 [0.655, 0.687] | -0.012 [-0.020, -0.002] | p=0.020 | 0.544 | 0.346 |
| Qwen3-0.6B | precision disagreement (dev-selected) [w6g128, pdrop] | 0.76 | 0.684 [0.666, 0.700] | +0.001 [-0.009, +0.011] | p=0.760 | 0.559 | 0.343 |
| Qwen3-0.6B | option order: 1 - p'[answer] | 1.00 | 0.732 [0.716, 0.748] | +0.049 [+0.031, +0.067] | p=0.003 | 0.603 | 0.312 |
| Qwen3-0.6B | option order: flip | 1.00 | 0.657 [0.641, 0.671] | -0.026 [-0.046, -0.006] | p=0.007 | 0.556 | 0.327 |
| Qwen3-0.6B | option order: JSD | 1.00 | 0.650 [0.633, 0.666] | -0.032 [-0.056, -0.012] | p=0.007 | 0.554 | 0.375 |
| Qwen3-0.6B | logistic, confidence features | 0.00 | 0.682 [0.666, 0.699] | -0.000 [-0.002, +0.002] | p=0.840 | 0.557 | 0.341 |
| Qwen3-0.6B | logistic, confidence + precision | 0.76 | 0.691 [0.674, 0.707] | +0.009 [+0.002, +0.015] | p=0.020 | 0.563 | 0.337 |
| Qwen3-0.6B | logistic, confidence + option order | 1.00 | 0.742 [0.727, 0.758] | +0.060 [+0.047, +0.073] | p=0.003 | 0.601 | 0.303 |
| Qwen3-0.6B | logistic, all three | 1.76 | 0.743 [0.728, 0.759] | +0.061 [+0.048, +0.074] | p=0.003 | 0.606 | 0.303 |
| Qwen3-1.7B | entropy of 8-bit answer | 0.00 | 0.767 [0.752, 0.783] | - | - | 0.619 | 0.150 |
| Qwen3-1.7B | 1 - max probability | 0.00 | 0.766 [0.750, 0.782] | -0.002 [-0.003, -0.000] | p=0.020 | 0.615 | 0.151 |
| Qwen3-1.7B | negative margin | 0.00 | 0.763 [0.747, 0.779] | -0.004 [-0.007, -0.002] | p=0.003 | 0.610 | 0.152 |
| Qwen3-1.7B | precision disagreement (dev-selected) [w6g128, pdrop] | 0.76 | 0.766 [0.752, 0.779] | -0.001 [-0.009, +0.006] | p=0.713 | 0.627 | 0.155 |
| Qwen3-1.7B | option order: 1 - p'[answer] | 1.00 | 0.809 [0.796, 0.823] | +0.042 [+0.028, +0.056] | p=0.003 | 0.670 | 0.135 |
| Qwen3-1.7B | option order: flip | 1.00 | 0.705 [0.690, 0.721] | -0.062 [-0.077, -0.043] | p=0.003 | 0.638 | 0.165 |
| Qwen3-1.7B | option order: JSD | 1.00 | 0.797 [0.784, 0.809] | +0.029 [+0.016, +0.042] | p=0.003 | 0.643 | 0.139 |
| Qwen3-1.7B | logistic, confidence features | 0.00 | 0.765 [0.749, 0.781] | -0.003 [-0.004, -0.001] | p=0.007 | 0.614 | 0.151 |
| Qwen3-1.7B | logistic, confidence + precision | 0.76 | 0.770 [0.755, 0.785] | +0.003 [-0.001, +0.006] | p=0.087 | 0.626 | 0.150 |
| Qwen3-1.7B | logistic, confidence + option order | 1.00 | 0.815 [0.802, 0.829] | +0.048 [+0.038, +0.057] | p=0.003 | 0.678 | 0.131 |
| Qwen3-1.7B | logistic, all three | 1.76 | 0.817 [0.804, 0.831] | +0.050 [+0.040, +0.059] | p=0.003 | 0.681 | 0.131 |

## 6 Controls: bits, group size and matched noise

Figure 4 plots the AUROC of each disagreement score against the effective bits of the low-precision copy. For 1 - p_low[answer] the AUROC over all questions rises with bits over the range that was run, from 0.472 at 3 bits to 0.688 for w6g128, which is the best of the 9 copies; the entropy of the 8-bit copy gets 0.686 on the same questions. No copy lies clearly above the entropy line. The curve stops at the highest precision tried because the next step is the 8-bit copy itself, where the disagreement is dominated by rounding. The trade-off described in Section 4 is therefore visible only from one side: the signal is best where the quantization damage is small.

Figure 6 puts all 14 perturbed copies (rounding and noise) on one axis, the share of answers that differ from the 8-bit copy. The rank correlation between that damage and the AUROC is -0.895: the more a copy flips, the less its probability of the original answer says about whether that answer is right. Rounding and noise points interleave along the same trend.

At 4 bits, group sizes 32, 64, 128 and 256 give 0.587, 0.552, 0.550 and 0.584 (Appendix B). The ordering is not monotone in group size, which is what one expects if the AUROC mostly follows how much damage a configuration does and the damage of min-max rounding is irregular across group sizes.

Table 3 and Figure 5 are the control that decides whether the signal belongs to quantization. At the error energy of the 4-bit group-128 quantizer the AUROC of 1 - p_low[answer] is 0.550 for rounding. It is 0.597 (range 0.531 to 0.649, 3 seeds) for Gaussian noise with a per-tensor variance and 0.511 (range 0.468 to 0.553, 2 seeds) for noise with a per-group variance. The noisy copies are about as informative as the rounded copy, and the seed-to-seed spread of the noise is as large as the gap between conditions, so the signal is not specific to the quantizer. Matching the error energy does not match the damage, though. The five noise draws flip between 45% and 68% of the answers against 51% for rounding, and across the five draws the AUROC falls as the flips rise (rank correlation -0.900). The rounded copy is one deterministic draw, so its seed spread is unknown, and the intervals of Table 3 cover the seeds that were run only. This matches a picture in which the copies are samples from a cloud around the original weights and only the size of the cloud matters. The 4-bit cloud is too large for a 0.6B model, as Section 4 showed.

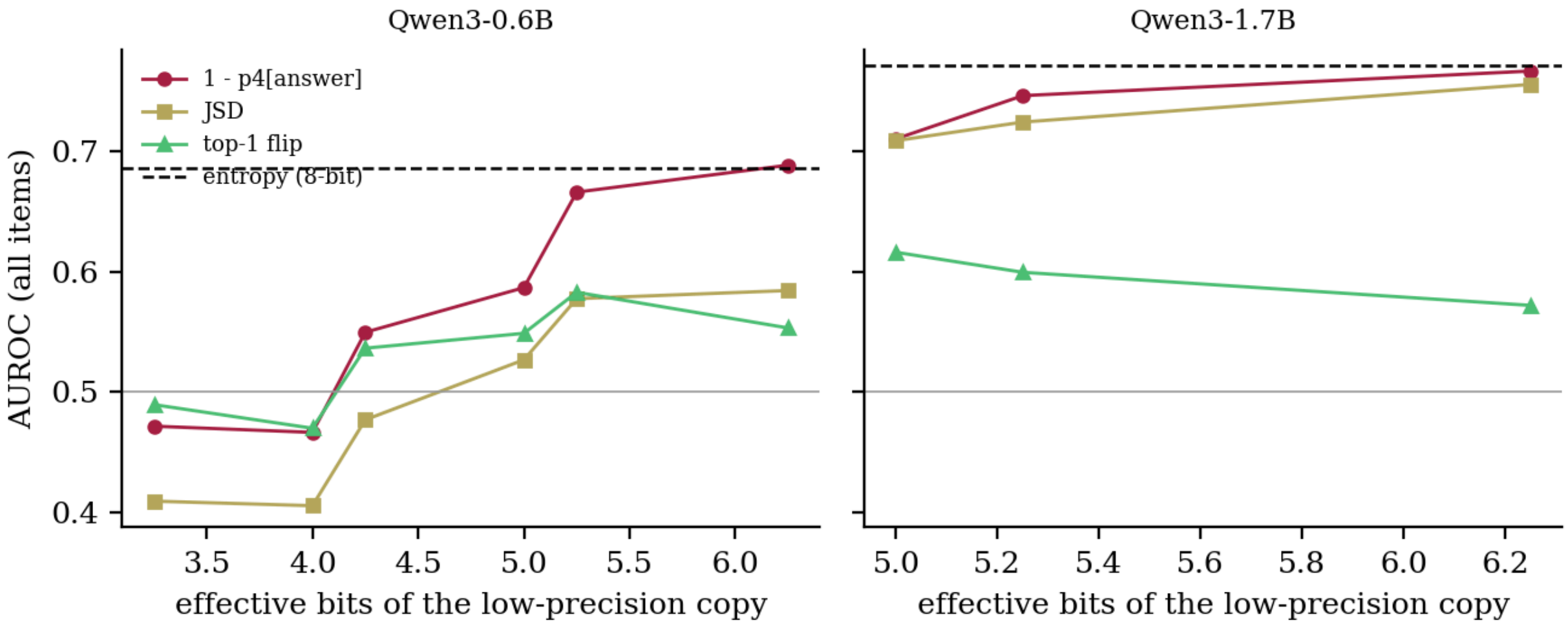

**Figure 4.** AUROC against the effective bits of the low-precision copy for the three disagreement scores on all questions. The dashed line is the entropy of the 8-bit copy. Too little perturbation flips nothing and too much replaces the model with a different, worse one.

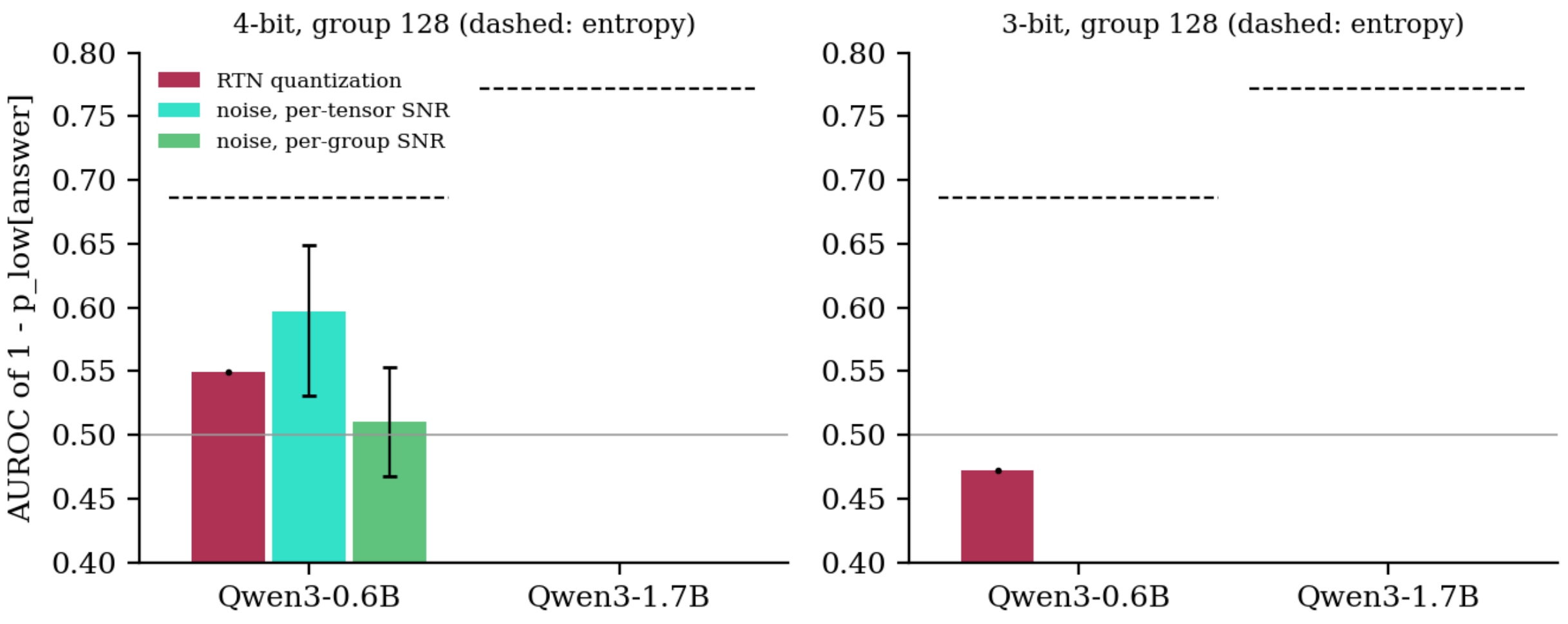

**Figure 5.** Matched-noise control. At the same error energy as the 4-bit group-128 quantizer (left), Gaussian weight noise of the same size is added to the 16-bit weights; the 3-bit panel (right) shows rounding only, because no 3-bit noise copies were run, and no noise copies were run for the 1.7B model. Bars: AUROC of 1 - p_low[answer]; error bars span the noise seeds that were run; dashed line: entropy.

**Table 3.** Matched-noise control (group 128). Accuracy and flip rate of the perturbed copy, and AUROC of 1 - p_low[answer] and of the JSD, as mean over noise seeds with the range in brackets. Noise has the same error energy as the quantizer.

| model | error energy of | perturbation | seeds | copy acc. | flips vs 8-bit | AUROC 1 - p_low[ans] | AUROC JSD |
|---|---|---|---|---|---|---|---|
| Qwen3-0.6B | 4-bit | RTN quantization | 1 | 36.8% | 50.6% | 0.550 | 0.477 |
| Qwen3-0.6B | 4-bit | noise, per-tensor variance | 3 | 38.4% | 52.1% | 0.597 (0.531 to 0.649) | 0.533 (0.453 to 0.586) |
| Qwen3-0.6B | 4-bit | noise, per-group variance | 2 | 30.6% | 61.0% | 0.511 (0.468 to 0.553) | 0.448 (0.426 to 0.470) |
| Qwen3-0.6B | 3-bit | RTN quantization | 1 | 24.6% | 74.3% | 0.472 | 0.410 |

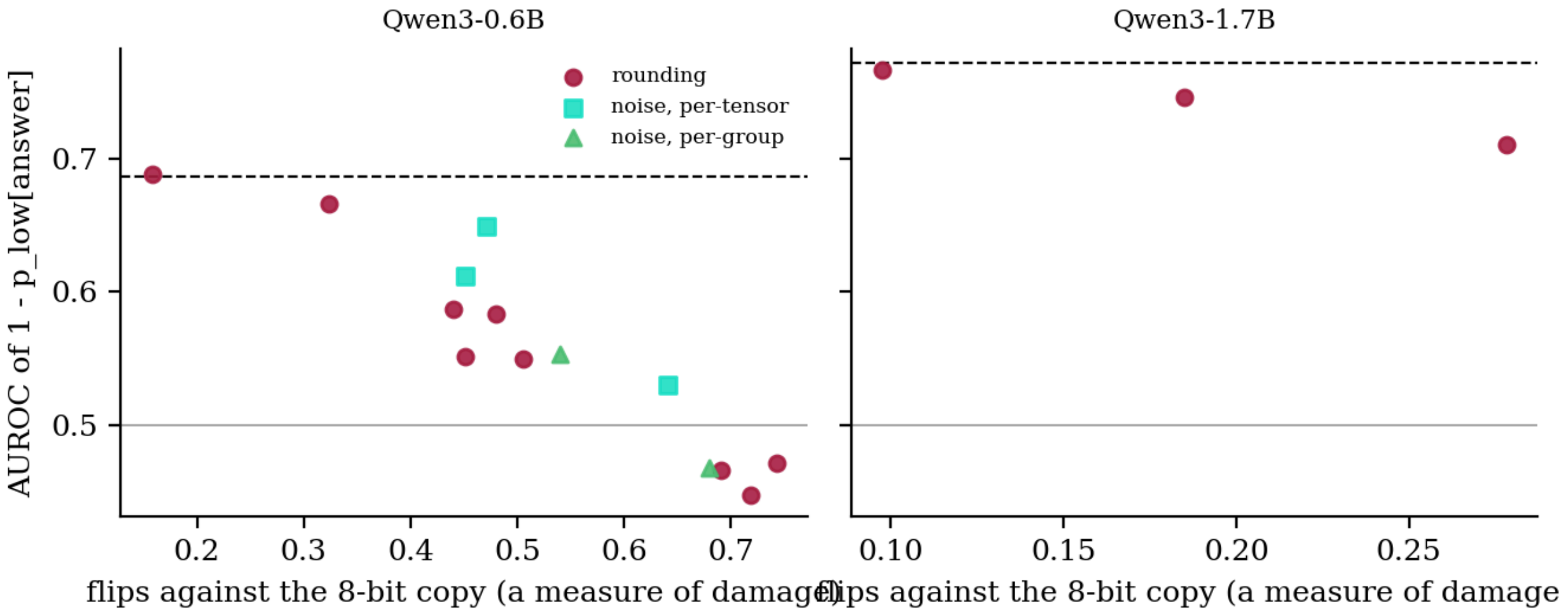

**Figure 6.** All perturbed copies on one axis: AUROC of 1 - p_low[answer] against the share of answers that differ from the 8-bit copy. Rounding and Gaussian noise fall on the same trend, which points to the amount of damage as what sets the quality of the signal. Dashed: entropy.

## 7 Does it add information beyond confidence?

The question that matters in practice is whether the disagreement score adds anything to what the 8-bit copy already knows about itself, and two tests address it. The first fits a logistic model on the development half, once on the three confidence features alone and once on those plus the four disagreement features. It compares the two on the test half with a paired bootstrap (Table 2). For the 0.6B model the gain is +0.009 [+0.002, +0.015] AUROC (p=0.020), and the partial AUROC at low false-alarm rates moves from 0.557 to 0.563. The effect is small, and one 50/50 split gives only one draw of the development fit. Two checks therefore use all 8,398 questions with the copy chosen on the development half. One is a likelihood-ratio test of the four added coefficients (chi-square 68 on 4 degrees of freedom, p<0.001). The other is the five-fold cross-fitted gain over the confidence features, +0.008 [+0.004, +0.013] for the 0.6B model. The same checks on the 1.7B model give chi-square 63 on 4 degrees of freedom, p<0.001, and a cross-fitted gain of +0.004 [+0.001, +0.006]. The 1.7B gain, which the test half alone could not separate from zero, is therefore also small and positive when all questions are used. The second test is by task (Table 4, Figure 7). On the 0.6B model the gain is positive with an interval above zero on ARC-Easy and ARC-Challenge, and inside the noise on MMLU subset and HellaSwag subset. On this model the tasks where it helps are the two with the highest accuracy of the 8-bit copy (66.8% and 49.8%). On the MMLU and HellaSwag subsets the 8-bit copy answers 41.8% and 39.6% correctly. A large part of the wrong answers are then guesses of a model that does not know, and neither the confidence nor the disagreement can tell them apart from lucky guesses. The 1.7B model does not follow that pattern. Its gain is +0.000 [-0.003, +0.003] on ARC-Easy, +0.000 [-0.006, +0.006] on ARC-Challenge, -0.001 [-0.006, +0.003] on the MMLU subset and +0.018 [+0.007, +0.028] on the HellaSwag subset, where it answers 60.7% correctly. Concentration on ARC is therefore a property of the 0.6B model in this study, and the accuracy explanation above is not supported by the larger model. The disagreement score alone also differs from entropy in opposite directions by task. ARC-Easy gives +0.019 [+0.007, +0.033] for the 0.6B model and -0.016 [-0.028, -0.004] for the 1.7B model. ARC-Challenge gives +0.019 [-0.001, +0.037] for the 0.6B model and -0.027 [-0.043, -0.012] for the 1.7B model. MMLU subset gives +0.006 [-0.006, +0.018] for the 0.6B model and -0.015 [-0.025, -0.003] for the 1.7B model. HellaSwag subset gives -0.008 [-0.022, +0.007] for the 0.6B model and +0.025 [+0.009, +0.039] for the 1.7B model. The pooled AUROC of the two scores does not show these within-task differences, because it also ranks questions across tasks.

Figure 9 cuts the questions by the entropy of the 8-bit answer. In the most confident third the error rate is 29.5%; there entropy itself reaches an AUROC of 0.673, the precision disagreement 0.658, and the option-order score 0.755. Inside the middle and least confident thirds, where the error rates are 56.2% and 65.8%, the entropy drops to 0.551 and 0.537 and the precision disagreement reaches 0.580 and 0.550. The scatter in Figure 10 shows why precision disagreement adds little: its rank correlation with entropy is 0.832, so most of what it says, the confidence already said. The option-order score has a correlation of 0.472 with entropy, so it carries information from a different place, which is consistent with its larger gain.

To check whether the gain is specific to rounding, the same cross-fitted test was repeated for every perturbed copy of the 0.6B model (Figure 8). The gain over the confidence features has an interval above zero for 4 of 9 rounded copies and 1 of 5 noisy copies. The largest rounded gain is +0.010 [+0.006, +0.015] (w5g128) and the largest noisy gain is +0.015 [+0.009, +0.020].

A control in which the four disagreement features of w6g128 are shuffled across questions, with the same model and the same number of features, gives +0.000 [-0.001, +0.002]. A stricter control shuffles them only among questions of the same task and the same tenth of the entropy. Any link to task and to confidence is then kept and only the question-specific part is destroyed. Over 20 shuffles the gain is -0.0005 on average (largest +0.0012) against the observed +0.0082. The extra columns alone therefore do not produce the gain, and it comes from question-level information in the second reading. Rounding and random noise share it, which is consistent with a gain that comes from reading the same weights twice under perturbation. With 14 intervals in Figure 8 and none corrected for multiplicity, the count above is descriptive. 3 of the 3 rounded copies of the 1.7B model have an interval above zero in the same analysis (gains from +0.004 to +0.007). Copies that flip about half of the answers mostly add little, in line with the damage trend of Figure 6. The spread between noise draws of one energy (gains from -0.000 to +0.015) shows how much a single perturbation draw matters.

**Table 4.** Per-task AUROC over all questions of the task: entropy, the disagreement score alone, and the 5-fold cross-fitted logistic models without and with the disagreement features. The last columns give the paired difference between the two logistic models.

| model | task | n | acc. | entropy | disagr. alone | conf. only (logit) | conf. + disagr. | difference [95% CI] | p |
|---|---|---|---|---|---|---|---|---|---|
| Qwen3-0.6B | ARC-Easy | 2,376 | 66.8% | 0.757 | 0.777 | 0.756 | 0.776 | +0.020 [+0.011, +0.027] | p=0.003 |
| Qwen3-0.6B | ARC-Challenge | 1,172 | 49.8% | 0.681 | 0.700 | 0.678 | 0.702 | +0.024 [+0.012, +0.035] | p=0.003 |
| Qwen3-0.6B | MMLU subset | 2,850 | 41.8% | 0.659 | 0.665 | 0.659 | 0.666 | +0.007 [-0.002, +0.016] | p=0.147 |
| Qwen3-0.6B | HellaSwag subset | 2,000 | 39.6% | 0.568 | 0.560 | 0.567 | 0.561 | -0.006 [-0.015, +0.004] | p=0.233 |
| Qwen3-1.7B | ARC-Easy | 2,376 | 86.6% | 0.839 | 0.823 | 0.837 | 0.837 | +0.000 [-0.003, +0.003] | p=0.920 |
| Qwen3-1.7B | ARC-Challenge | 1,172 | 71.7% | 0.793 | 0.766 | 0.788 | 0.788 | +0.000 [-0.006, +0.006] | p=0.900 |
| Qwen3-1.7B | MMLU subset | 2,850 | 54.6% | 0.728 | 0.712 | 0.722 | 0.720 | -0.001 [-0.006, +0.003] | p=0.587 |
| Qwen3-1.7B | HellaSwag subset | 2,000 | 60.7% | 0.677 | 0.702 | 0.672 | 0.690 | +0.018 [+0.007, +0.028] | p=0.003 |

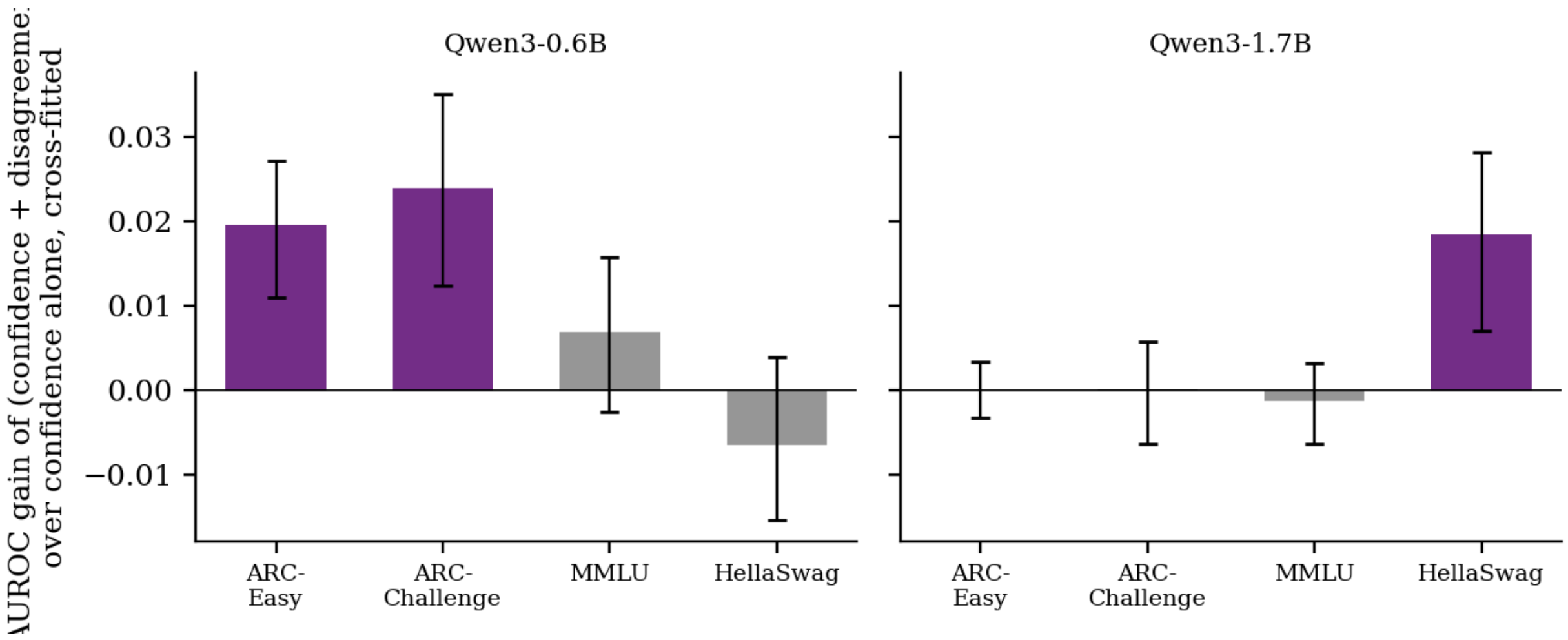

**Figure 7.** AUROC gain of the logistic combination of confidence and precision disagreement over confidence alone, per task, 5-fold cross-fit, with paired bootstrap intervals.

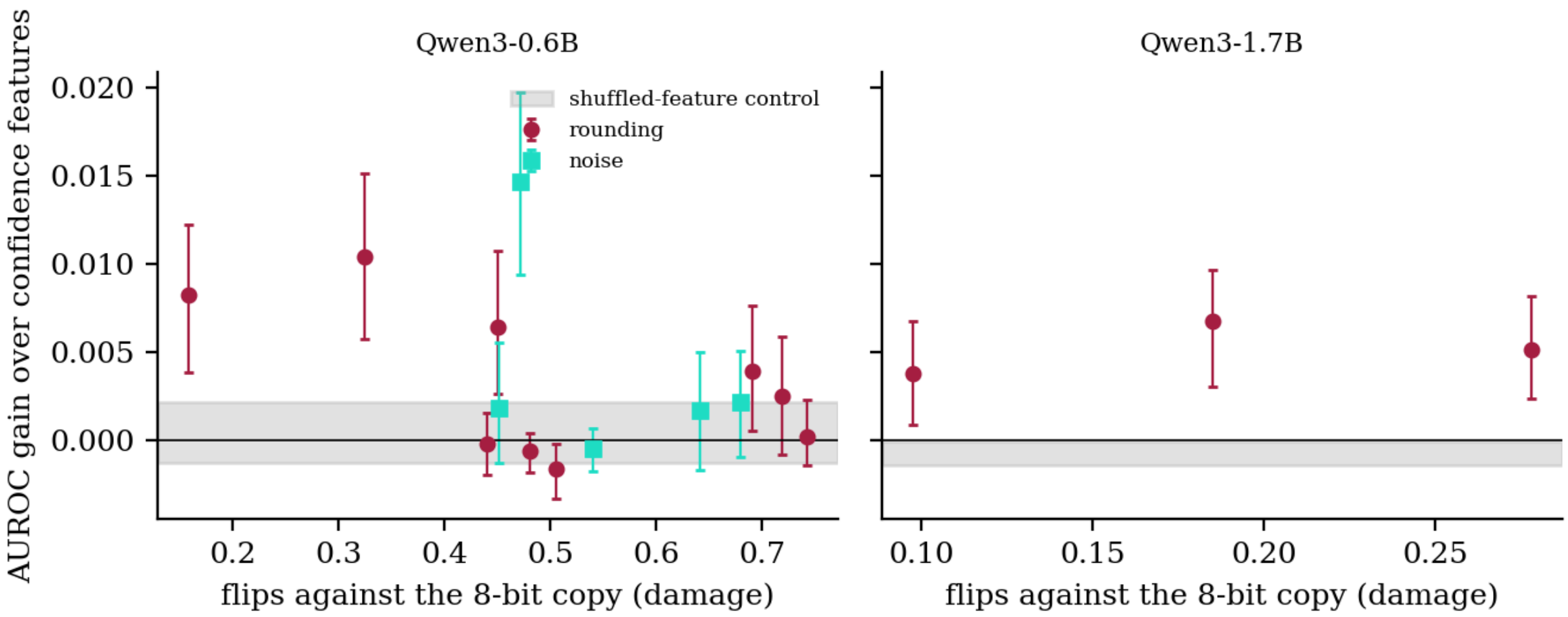

**Figure 8.** Added information beyond the confidence features, copy by copy. For each rounded or noisy copy, the AUROC gain from adding its four disagreement features to the three confidence features (5-fold cross-fit, paired bootstrap intervals). The grey band is the interval of a control in which the disagreement features of the selected copy are shuffled across questions.

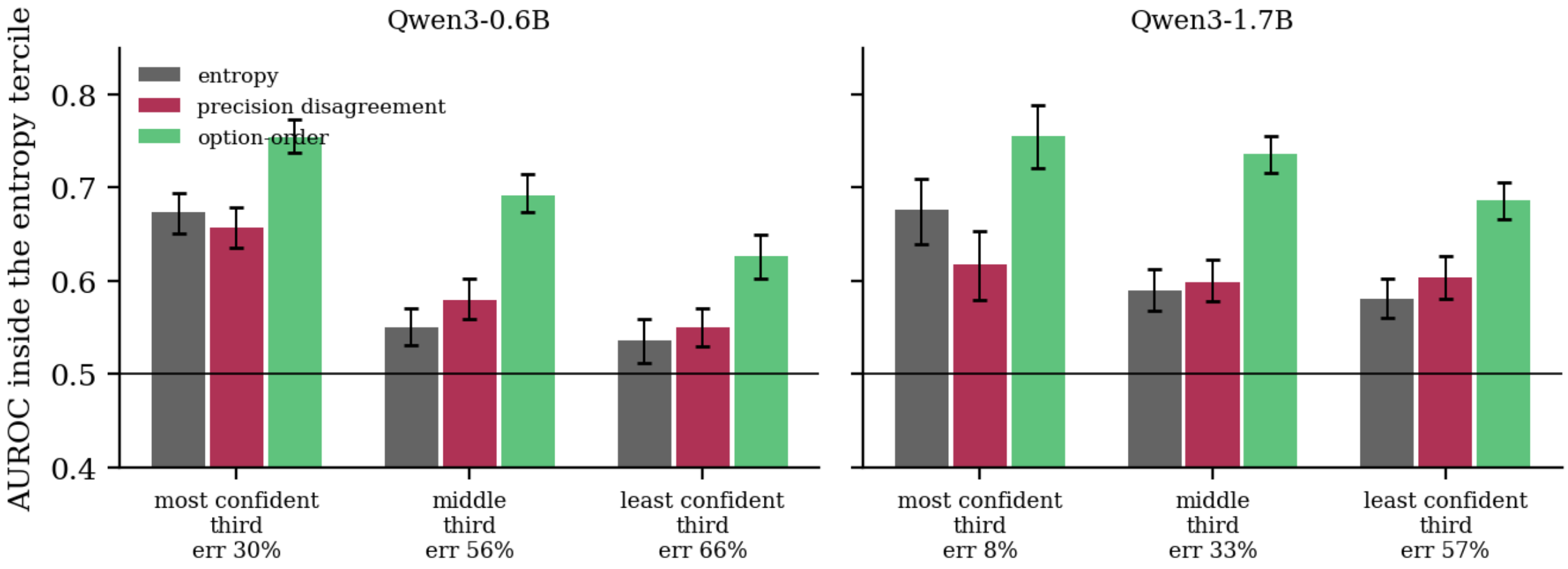

**Figure 9.** AUROC inside terciles of the 8-bit entropy, with bootstrap intervals, to show whether a score helps where confidence is already high. Error rates of the terciles are under the labels.

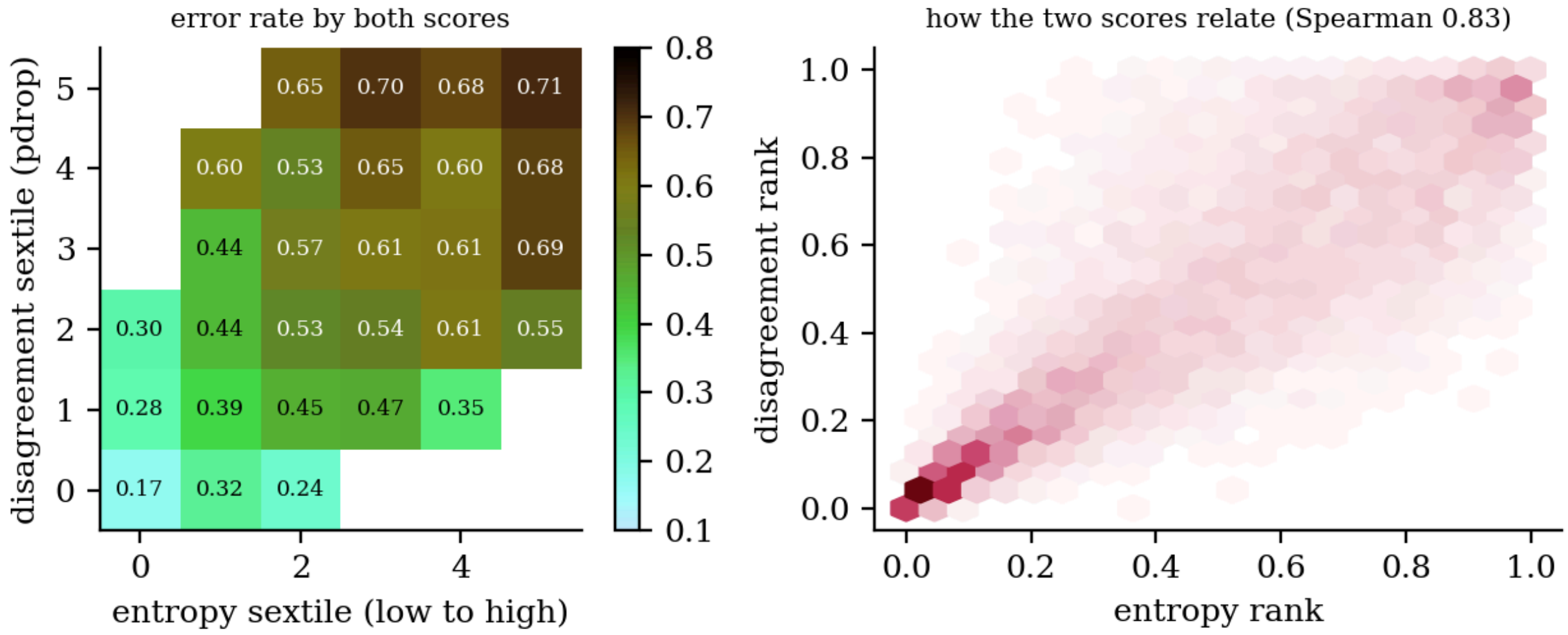

**Figure 10.** Left: error rate of the 0.6B model as a function of both scores, in sextiles of each. Right: how the two scores relate. If the disagreement score only repeated the entropy the left panel would be constant along its vertical axis.

## 8 Selective prediction and calibration

Figure 11 and Table 5 turn the scores into selective-prediction curves. For the 0.6B model the test-half accuracy without abstention is 49.6%. Answering only the 80% of questions the entropy ranks as most certain lifts it to 53.7%; with the precision-disagreement score it is 53.8% and with the option-order score 55.9%. At 50% coverage the figures are 62.3%, 63.0% and 66.2%. The AURC is 0.341 for entropy and 0.343 for the precision disagreement, intervals that overlap almost completely; the option-order score is lower at 0.312. Combining all three in a logistic model gives 67.4% at 50% coverage. The ranking of the scores in these curves is the ranking of their AUROCs, with no surprise at the high-coverage end that would change the picture.

The raw max-probability of the small model is overconfident: its ECE on the test half is 0.181. A one-feature logistic recalibration of the entropy, fitted on the development half, brings it to 0.032, and the three confidence features together to 0.014 (Table 6, Figure 12). Adding the precision-disagreement features gives 0.020 and the option-order features 0.025. After recalibration the 0.6B model's confidences are within 0.014 to 0.032 ECE of perfect calibration, and the intervals of the recalibrated variants overlap. The 1.7B model is different: the entropy-only map leaves 0.073 [0.063, 0.087], while the three confidence features give 0.021 [0.016, 0.037] and the precision-disagreement features 0.018, so a one-feature map is not enough there. The benefit of a better score shows up as sharper confidences and not as a lower ECE: the Brier score falls from 0.221 to 0.204 with the option-order features. A simple recalibration is worth more than any of the disagreement scores for calibration, and it costs nothing at inference time.

The unpaired AURC intervals of Table 5 overlap, so we also give paired differences from the entropy AURC (negative is better). For the 0.6B model the precision-disagreement score gives +0.002 [-0.003, +0.007] and the logistic combination with confidence -0.004 [-0.007, -0.001]. For the 1.7B model the same two give +0.005 [+0.001, +0.009] and +0.000 [-0.001, +0.001]. On the larger model the single disagreement score therefore has a worse AURC than entropy.

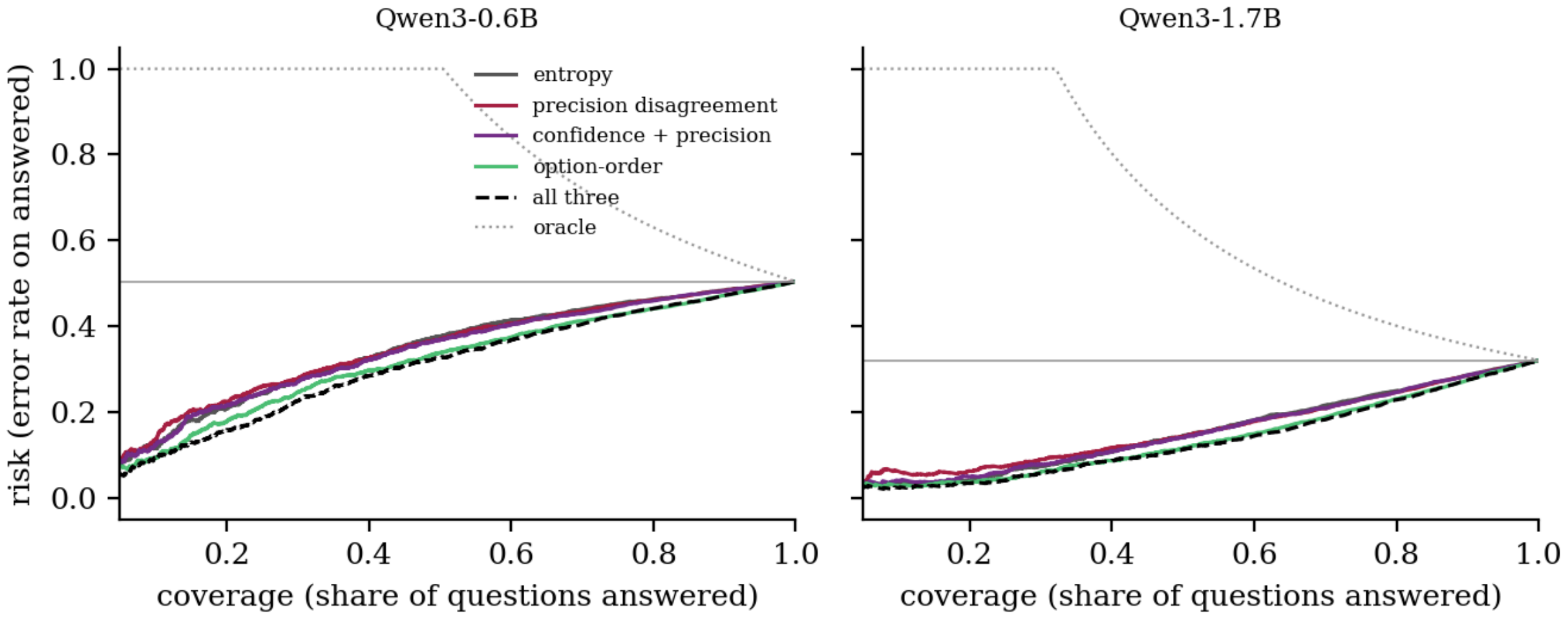

**Figure 11.** Risk-coverage curves on the test half: error rate among the answered questions as the model answers fewer. Lower is better; the dotted curve is an oracle that abstains on exactly the wrong answers.

**Table 5.** Selective prediction on the test half. AURC (lower is better) and the accuracy on the answered questions at 50% and 80% coverage, with 95% bootstrap intervals over questions. Accuracy without abstention is 49.6% for the 0.6B model, 67.9% for the 1.7B model.

| model | score | AURC [95% CI] | acc. at 50% coverage | acc. at 80% coverage |
|---|---|---|---|---|
| Qwen3-0.6B | entropy | 0.341 [0.324, 0.359] | 0.623 [0.602, 0.642] | 0.537 [0.523, 0.553] |
| Qwen3-0.6B | precision disagreement | 0.343 [0.327, 0.362] | 0.630 [0.608, 0.650] | 0.538 [0.522, 0.555] |
| Qwen3-0.6B | confidence + precision (logit) | 0.337 [0.320, 0.354] | 0.632 [0.609, 0.652] | 0.540 [0.525, 0.557] |
| Qwen3-0.6B | option order | 0.312 [0.297, 0.328] | 0.662 [0.643, 0.683] | 0.559 [0.542, 0.578] |
| Qwen3-0.6B | all three (logit) | 0.303 [0.288, 0.319] | 0.674 [0.651, 0.692] | 0.558 [0.542, 0.576] |

| model | score | AURC [95% CI] | acc. at 50% coverage | acc. at 80% coverage |
|---|---|---|---|---|
| Qwen3-1.7B | entropy | 0.150 [0.138, 0.161] | 0.855 [0.840, 0.870] | 0.750 [0.737, 0.765] |
| Qwen3-1.7B | precision disagreement | 0.155 [0.143, 0.167] | 0.857 [0.843, 0.871] | 0.754 [0.739, 0.770] |
| Qwen3-1.7B | confidence + precision (logit) | 0.150 [0.138, 0.162] | 0.859 [0.846, 0.874] | 0.753 [0.738, 0.769] |
| Qwen3-1.7B | option order | 0.135 [0.124, 0.146] | 0.884 [0.870, 0.896] | 0.769 [0.754, 0.783] |
| Qwen3-1.7B | all three (logit) | 0.131 [0.120, 0.140] | 0.886 [0.874, 0.900] | 0.771 [0.759, 0.785] |

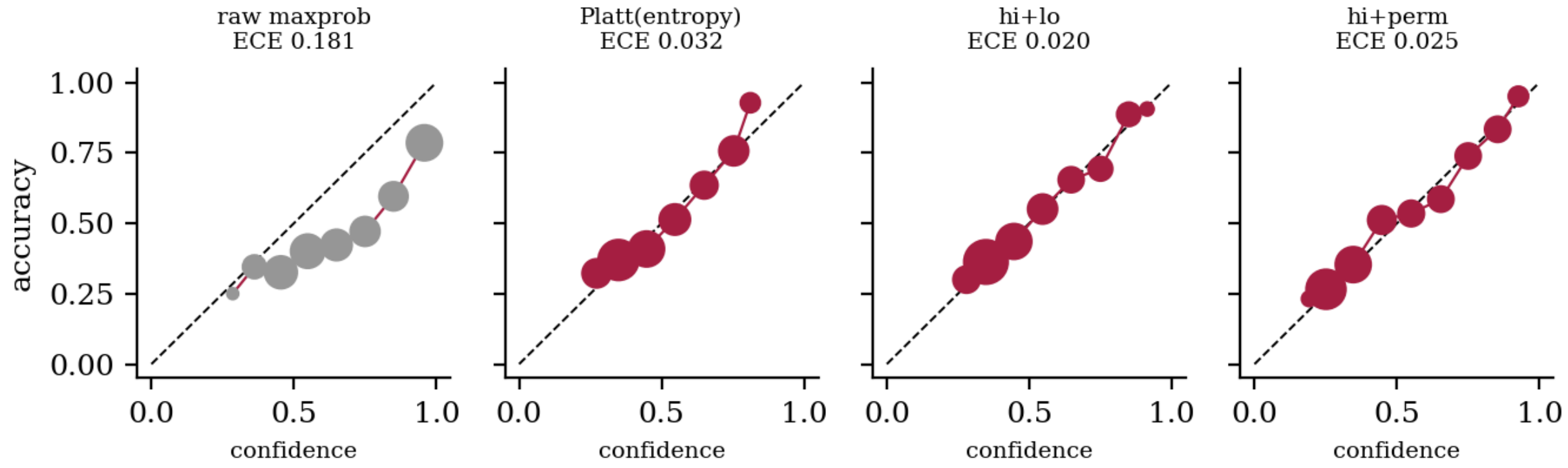

**Figure 12.** Reliability of the confidence after recalibration on the development half, shown on the test half for the 0.6B model. Point area grows with the number of questions in the bin; ECE in each title.

**Table 6.** Calibration on the test half. The confidence is the probability of being correct from a logistic map fitted on the development half (the raw max-probability row is uncalibrated). ECE with 15 bins and Brier score.

| model | confidence | ECE [95% CI] | Brier |
|---|---|---|---|
| Qwen3-0.6B | raw max probability | 0.181 [0.168, 0.194] | 0.260 |
| Qwen3-0.6B | logistic map of entropy | 0.032 [0.025, 0.050] | 0.222 |
| Qwen3-0.6B | logistic, confidence features | 0.014 [0.013, 0.033] | 0.221 |
| Qwen3-0.6B | logistic, confidence + precision | 0.020 [0.016, 0.037] | 0.220 |
| Qwen3-0.6B | logistic, confidence + option order | 0.025 [0.021, 0.042] | 0.204 |
| Qwen3-0.6B | logistic, all three | 0.027 [0.022, 0.043] | 0.204 |
| Qwen3-1.7B | raw max probability | 0.212 [0.197, 0.225] | 0.235 |
| Qwen3-1.7B | logistic map of entropy | 0.073 [0.063, 0.087] | 0.186 |
| Qwen3-1.7B | logistic, confidence features | 0.021 [0.016, 0.037] | 0.178 |
| Qwen3-1.7B | logistic, confidence + precision | 0.018 [0.017, 0.035] | 0.176 |
| Qwen3-1.7B | logistic, confidence + option order | 0.022 [0.018, 0.038] | 0.160 |
| Qwen3-1.7B | logistic, all three | 0.019 [0.016, 0.035] | 0.159 |

## 9 Scale, tasks and domain shift

Table 7 and Figure 13 put the two model sizes side by side. The 1.7B model answers 67.5% of the questions correctly against 49.5% for the 0.6B model, and the entropy of its 8-bit answer detects errors with AUROC 0.767 against 0.682. The selected low-precision copy is w6g128 for the 0.6B model (among 9 copies) and w6g128 for the 1.7B model (among 3); in both cases it is the highest-precision copy that was run. The logistic combination with precision disagreement differs from entropy by +0.003 [-0.001, +0.006] on the 1.7B model, and the option-order score by +0.042 [+0.028, +0.056]. The larger model tolerates quantization better, so a given bit-width does less damage and the useful range of the low-precision copy moves down (Table 1).

Figure 14 fits the combination on one domain and tests it on another: four MMLU categories (STEM, humanities, social sciences, other), ARC-Challenge and HellaSwag. Across the 30 off-diagonal source-target pairs for the 0.6B model, the confidence-only logistic model has a mean AUROC of 0.643 and adding the disagreement features gives 0.648. The change is +0.005 on average, between -0.011 and +0.023, and positive in 23 of 30 pairs. The paired bootstrap interval of the change is above zero for 4 pairs and below zero for 0. The pairs share questions, so they are not 30 independent tests. The mean ECE after transfer changes by +0.0030, which is slightly worse calibration, since a higher ECE is worse. In-domain (the diagonal, two-fold cross-fit) the mean AUROC is 0.637. That number is not a like-for-like reference. Each diagonal model is fitted on half of its own domain only, while the off-diagonal models use the whole source domain, so we draw no conclusion about the cost of a shift from it. Adding the disagreement features neither helped nor hurt measurably under the shifts tested. The conclusion is limited to the sort of shift seen here, where all domains are multiple-choice knowledge or commonsense questions with the same prompt format and the same models. Per-subject results vary widely around the pooled value with only 50 questions per subject, so we do not read individual subjects.

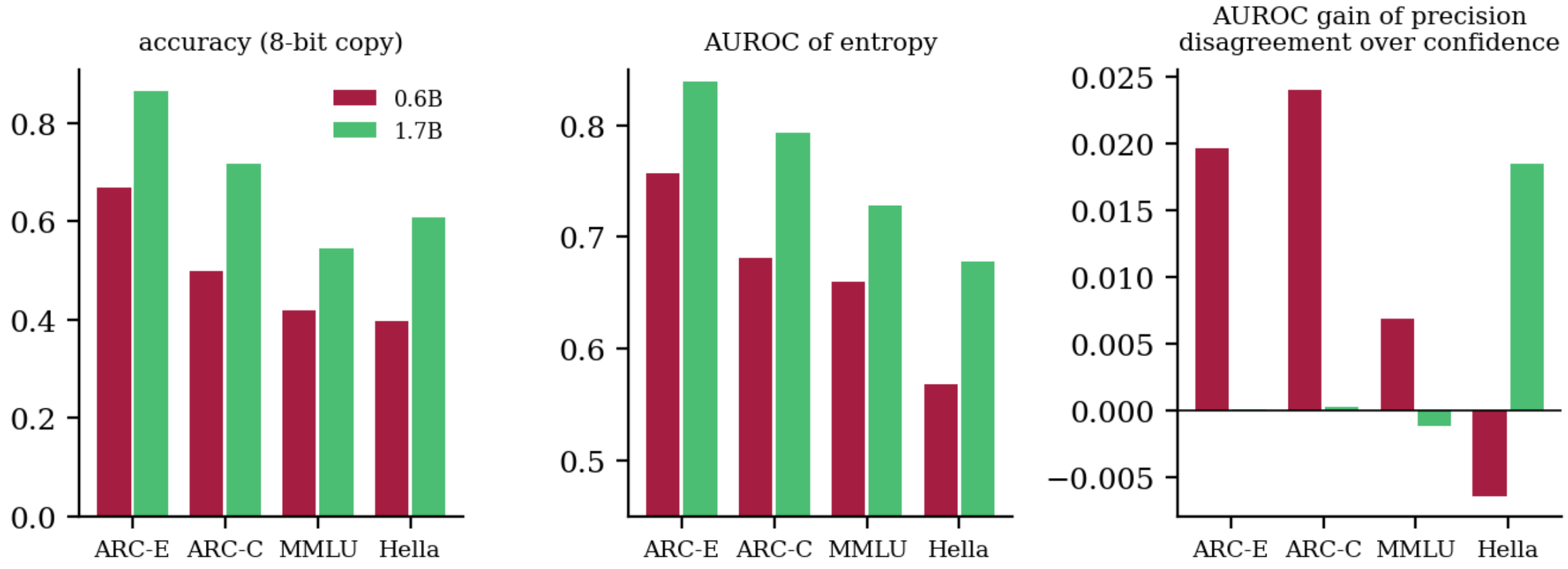

**Figure 13.** Model scale. Left: accuracy of the 8-bit copy. Middle: AUROC of entropy. Right: gain in AUROC when precision disagreement is added to confidence (cross-fitted).

**Table 7.** Two model sizes on the same questions (test half). Selected configuration and score, AUROC of entropy and of the selected disagreement score, and the paired differences from entropy for the logistic combination with precision disagreement and for the option-order score.

| model | acc. (8-bit) | selected copy / score | entropy AUROC | disagr. AUROC | conf.+precision minus entropy | option order minus entropy |
|---|---|---|---|---|---|---|
| Qwen3-0.6B | 49.5% | w6g128 / pdrop | 0.682 | 0.684 | +0.009 [+0.002, +0.015] | +0.049 [+0.031, +0.067] |
| Qwen3-1.7B | 67.5% | w6g128 / pdrop | 0.767 | 0.766 | +0.003 [-0.001, +0.006] | +0.042 [+0.028, +0.056] |

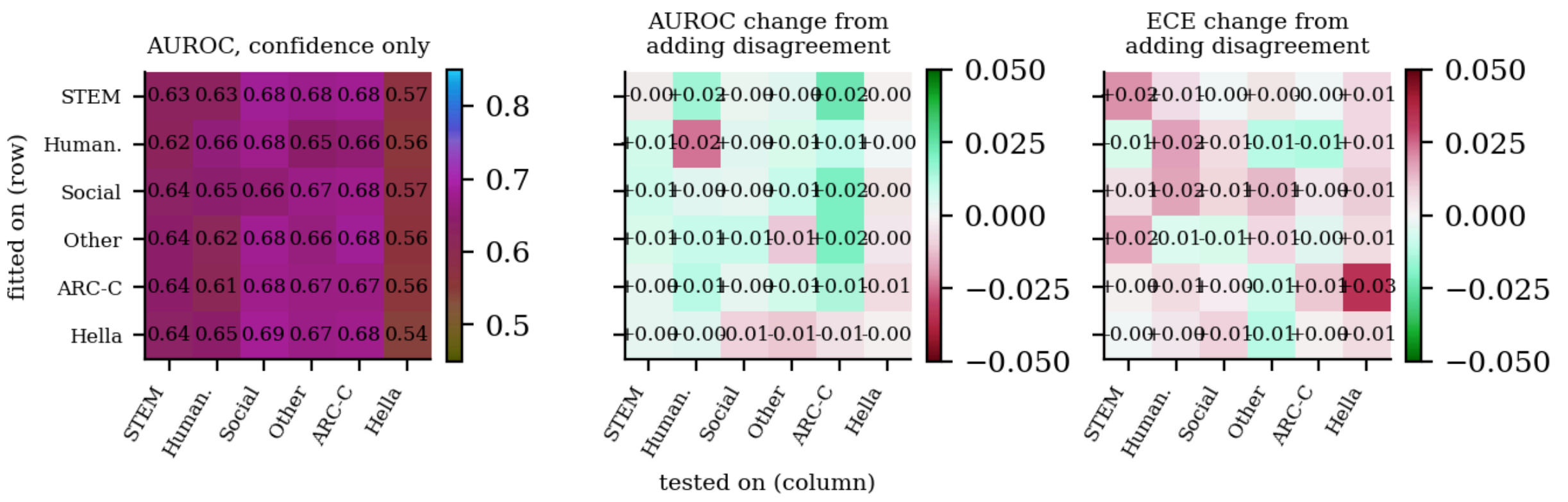

**Figure 14.** Shift test with the 0.6B model: the logistic combination is fitted on one domain (row) and tested on another (column); MMLU categories plus ARC-Challenge and HellaSwag. Left: AUROC with confidence features only. Middle and right: change in AUROC and in ECE from adding the disagreement features.

## 10 Cost comparison

Table 8 and Figure 15 line the scores up by what they cost. The free entropy has AUROC 0.686, and one extra low-precision pass costs 0.76 of an 8-bit pass in weight traffic and gives 0.688. One extra option-order pass at 8-bit precision costs 1.00 and gives 0.737; it reads about a third more weight bytes than the low-precision pass. The comparison is therefore close to equal cost but not exact, and a pass of the low-precision copy with shifted options, which would be at equal cost, was not run. Two low-precision copies together cost 1.39 and give 0.690. Spending more does not help in a regular way: four 4-bit copies voting reach 0.574 for 2.17 units, because each member is individually far from the original. The mutual information of those copies is at chance (0.500). Three additional option orders at 8-bit precision reach 0.735 for the vote and 0.739 for the entropy of the average, at three units. Averaging over the four option orders also changes the answer: its accuracy is 55.4% against 49.5% for the single order. Self-consistency with sampled reasoning chains was not completed within the compute budget, and a pilot on a few dozen questions is too small to report. It needs hundreds of decoded tokens per question. It would cost far more weight traffic than any of the passes in the table, so it is left out of the comparison.

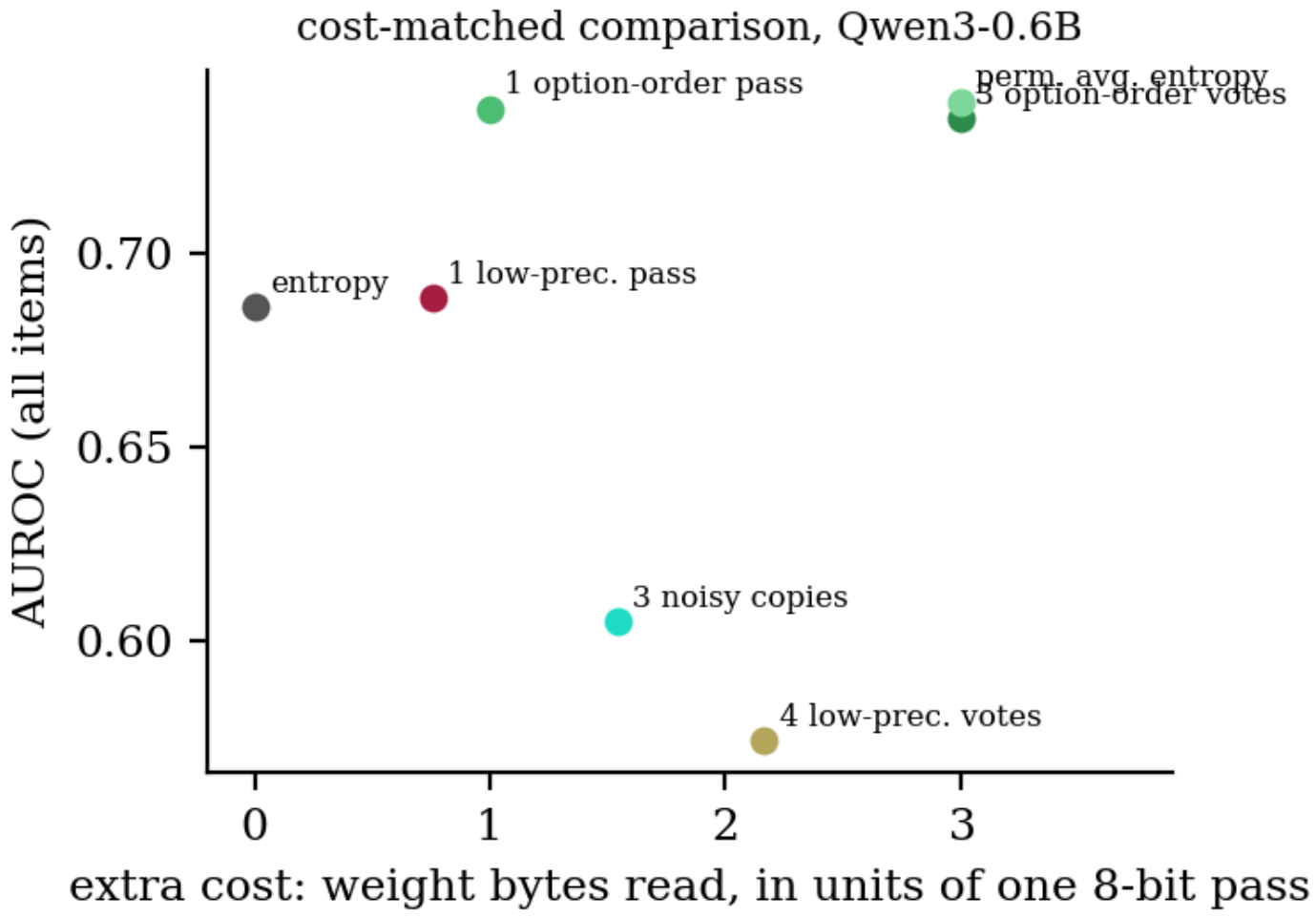

**Figure 15.** Cost comparison. AUROC against the extra weight bytes read, in units of one 8-bit pass.

**Table 8.** Cost comparison on all questions of the 0.6B model. Cost is the extra weight traffic per question in units of one 8-bit pass. 'Cheap' passes read fewer bytes than an 8-bit pass.

| score | extra passes | extra cost (8-bit pass units) | AUROC |
|---|---|---|---|
| entropy of the 8-bit answer | 0 | 0 | 0.686 |
| one low-precision pass (w6g128), pdrop | 1 (cheap) | 0.76 | 0.688 |

| score | extra passes | extra cost (8-bit pass units) | AUROC |
|---|---|---|---|
| one option-order pass (8-bit), 1 - p'[answer] | 1 | 1.00 | 0.737 |
| two low-precision copies (w6g128, w5g128): mean 1 - p_low[answer] | 2 (cheap) | 1.39 | 0.690 |
| four 4-bit copies (groups 32, 64, 128, 256): share disagreeing | 4 (cheap) | 2.17 | 0.574 |
| same copies: mutual information of the ensemble | 4 (cheap) | 2.17 | 0.500 |
| three noisy 4-bit-energy copies: share disagreeing | 3 (cheap) | 1.55 | 0.605 |
| three more option orders (8-bit): share disagreeing | 3 | 3.00 | 0.735 |
| entropy of the average over four option orders | 3 | 3.00 | 0.739 |

## 11 Deployment recipe: abstain, escalate or ask again

With the 0.6B model as the front and the 1.7B model as the fallback (Table 9, Figure 16), answering the most uncertain 20% of questions with the larger model takes the system from 49.5% to 53.9% when entropy picks the questions. The precision-disagreement score gives 54.7%, the option-order score 56.7% and random selection 53.0%; the larger model alone gets 67.5%. At the same escalated share, precision disagreement is +0.008 [+0.002, +0.014] and option order +0.028 [+0.020, +0.036] in accuracy against entropy (paired bootstrap), so the scores do not route alike at equal share: option order routes best.

Equal share is not equal cost. Escalating 20% of the questions costs 0.2 × 3.2 = 0.64 units per question on average. That is less than the 0.76 of the low-precision pass and the 1.00 of the option-order pass, so the extra signal costs more than the escalation it steers. Option-order routing with 20% escalated uses 1.64 units beyond the first pass. At that traffic, free entropy routing can escalate 51% of the questions and reaches 61.3% against 56.7%. Precision-disagreement routing with 20% escalated uses 1.40 units. At that traffic entropy can escalate 44% and reaches 59.3% against 54.7%. The right panel of Figure 16 shows the same ordering over the whole range. With the larger model available, the extra pass is better spent on escalating more questions on the free entropy ranking.

Re-asking with the options shifted by one position and averaging the two probability vectors turns the 49.5% accuracy of the single ask into 52.3% when every question is re-asked. If only the most uncertain 20% are re-asked, the accuracy is 50.2% with entropy as the selector, 50.8% with precision disagreement and 52.0% with the option-order score; at 50% it is 51.6%, 52.1% and 52.3%. This is a different use of the second pass from error detection: the second reading is averaged into the answer. Counting cost, re-asking the 20% that entropy flags needs 0.20 extra units. The option-order and precision selectors need their full pass first (1.00 and 0.76). For option order that pass already holds the second reading of every question.

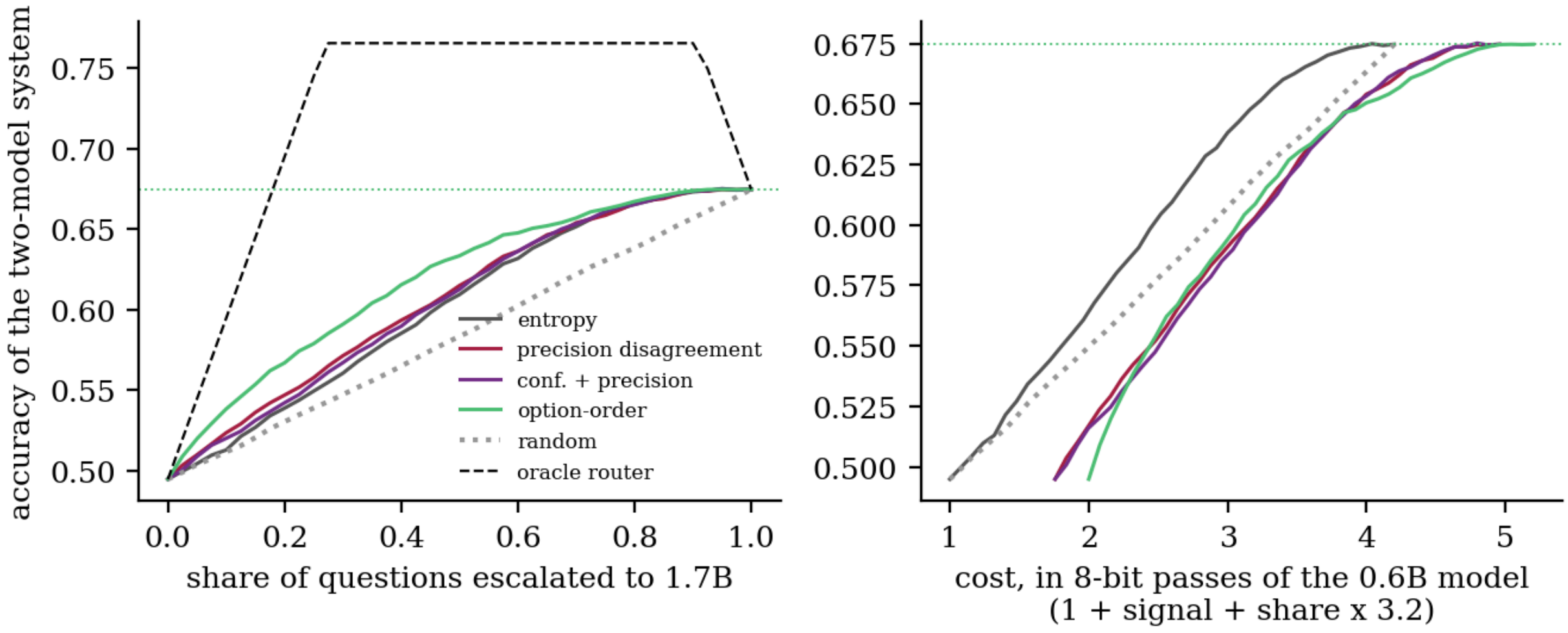

**Figure 16.** Escalation to the larger model. Accuracy of the two-model system when the most uncertain share of questions is answered by the larger model, against the share (left) and against the total weight bytes (right, in 8-bit passes of the small model).

**Table 9.** Escalation from the 0.6B model (49.5% correct) to the 1.7B model (67.5% correct) on all 8,398 questions. Accuracy of the two-model system when the most uncertain share of questions is answered by the larger model. The larger model costs 3.2 times the weight traffic of the smaller one per question.

| score used to route | signal cost | 10% escalated | 20% escalated | 30% escalated | 50% escalated |
|---|---|---|---|---|---|
| random | 0.00 | 51.3% | 53.0% | 54.8% | 58.4% |
| entropy | 0.00 | 51.3% | 53.9% | 56.1% | 60.9% |
| precision disagreement | 0.76 | 52.4% | 54.7% | 57.1% | 61.5% |
| confidence + precision (cross-fit) | 0.76 | 52.0% | 54.2% | 56.7% | 61.3% |
| option order | 1.00 | 53.8% | 56.7% | 59.1% | 63.3% |
| oracle router | - | 59.5% | 69.5% | 76.5% | 76.5% |

**Table 10.** What to do with which budget, as measured in this study. 'Free' means no extra forward pass.

| situation | recommended | why (this study) |
|---|---|---|
| no spare compute | abstain on a logistic map of the three confidence features, fitted on labelled questions (the study fitted on a development half of 4,199; smaller fitting sets were not evaluated) | free; ECE 0.014 (0.6B) and 0.021 (1.7B) after recalibration, against 0.181 raw (0.6B); the entropy-only map leaves 0.073 on the larger model |
| a spare low-precision copy is already in memory | add 1 - p_low[answer] as extra features to the recalibration; expect a small gain, of different size by model and task | paired AUROC gain over confidence +0.009 [+0.002, +0.015] (0.6B, test half); +0.018 [+0.007, +0.028] on the HellaSwag subset only for the larger model |
| one extra 8-bit pass is affordable and no larger model is available | re-ask with shifted options; use 1 - p'[answer] to flag questions, and average the two answers | AUROC 0.732; accuracy 49.5% to 52.3% when all are re-asked |
| a larger model is available | route the most uncertain 20 to 30% to it using the free entropy ranking | at equal total weight traffic entropy routing is ahead of both extra-pass scores (Figure 16); option order routes best only at equal share |
| need guarantees | do not rely on any of these scores alone | AUROC stays well below 1 for all of them on this model |

## 12 When it does not help

The data above already contain the cases where the method does not help. 3 of the 9 copies give a 1 - p_low[answer] AUROC below 0.5 or very close to it, and at 3 bits the JSD is anti-informative. The usable range is the narrow band of copies that still keep most of the accuracy, which is about 5 to 6 effective bits for the 0.6B model.

The task can also be too hard for the model. On the HellaSwag subset the combination gives -0.006 [-0.015, +0.004] over confidence and on MMLU +0.007 [-0.002, +0.016]. A model that is right 40% of the time on four-way questions has little signal of any kind to read. Difficulty alone does not predict where the method helps, though: the 1.7B model gains +0.018 [+0.007, +0.028] on the HellaSwag subset and nothing on ARC-Easy (+0.000 [-0.003, +0.003]), the reverse of the 0.6B model. A generic perturbation is as good as rounding: the matched-noise control (Table 3) shows that Gaussian weight noise of the same energy does as well, so a reader who does not already hold a quantized copy gains nothing from rounding specifically.

The disagreement signal also repeats the confidence. Its rank correlation with entropy is 0.832 (Figure 10), and the examples in Table 11 show how a false alarm arises: a question the 8-bit copy answers correctly with high confidence, where the damaged copy disagrees for reasons unrelated to the question. Ambiguous or mislabelled questions look uncertain to every score, and none of the scores can tell a hard question from a bad one.

**Table 11.** Worked examples from the test half of the 0.6B model, two drawn at random in each of three groups. Every question is in the most confident third by entropy; the groups differ in correctness and in whether the disagreement score is in its top fifth. p8, pL and p' are the probabilities given to the 8-bit answer by the 8-bit copy, by the w6g128 copy and by the 8-bit copy with shifted options.

| group | task | question (shortened) | gold | answer | p8 | pL | p' |
|---|---|---|---|---|---|---|---|
| caught by disagreement, missed by entropy | MMLU | Mill claims that the distinction between justice and other moral obligations corresponds perfectly to the dist... | A | D | 0.894 | 0.361 | 0.090 |
| caught by disagreement, missed by entropy | MMLU | What, according to Systemic theories, is the primary determinant of a state's foreign policy? | B | D | 0.855 | 0.238 | 0.321 |
| missed by both | HellaSwag | [header] How to fall in love [title] Boost your self-esteem. [step] To fall in love and find the right person ... | B | C | 0.903 | 0.893 | 0.115 |
| missed by both | MMLU | Which of the following is a characteristic of a developed country? | C | D | 0.972 | 0.961 | 0.294 |
| false alarm of disagreement | ARC-Easy | What happens to the water vapor in an air mass as it ascends? | B | B | 0.816 | 0.309 | 0.764 |
| false alarm of disagreement | ARC-Easy | Joey is making breakfast and pours a glass of milk over a bowl of dry cereal. Which best describes what Joey m... | C | C | 0.829 | 0.379 | 0.295 |

## 13 Limitations

The study covers two small models of one family, 0.6B and 1.7B parameters. A 4B confirmation run was not done because its bf16 weights alone exceed the 6 GB of GPU memory that this study was allowed to use. Larger models quantize with far less damage, so the useful low-precision range moves down and the picture at 8B or 70B may differ in either direction. Nothing here supports a claim about them.

The quantizer is round-to-nearest with per-group scale and offset, simulated and written for this study. Real deployments use GPTQ, AWQ, K-quants, imatrix or learned codebooks; those lose less accuracy at the same bit count and could change the balance between “informative” and “destroyed”. Bytes are a proxy for cost, since a real 4-bit kernel may be memory-bound or compute-bound and its speed depends on the hardware, so we report no timing. The cost units count the quantized projections only. The embedding table doubles as the output head and stays in 16 bits. It holds about 156 M weights in the 0.6B model. Counting it, the w6g128 pass costs 0.86 of an 8-bit pass instead of 0.76, which narrows the cost gap to the option-order pass.

The tasks are four-way multiple choice (a few ARC questions have three or five options), scored by letter logits with a zero-shot plain prompt. Accuracy of the small models is modest (42% on the MMLU subset for the 0.6B model), which limits how much any score can separate right from wrong, and the prompt format affects the base accuracy. Generative answers, long outputs and open-ended questions are outside this study.

Choosing the low-precision configuration among 9 candidates for the 0.6B model (3 for the larger one) on a development half and reporting the other half is clean for the headline numbers. The chosen copy is the highest-precision one in both models, so a copy between 6.25 and 8.25 effective bits was never tried and the best precision may lie there. The cross-fitted analyses reuse that choice and so leak slightly. All bootstrap intervals resample questions, not seeds or models. They do not cover variation across prompt templates, noise seeds beyond those listed, or checkpoints, and p-values are not corrected for the many comparisons in the tables. The matched-noise control was run for the 0.6B model only, at 4-bit energy, with 3 and 2 seeds, and rounding is a single deterministic copy. Noise-seed spread is shown where it was measured.

A question whose benchmark label is debatable will look uncertain to every score and is counted as an error for all of them, so the AUROC ceiling is below 1. The ambiguity studies cited in the related work suggest that all of these scores would look worse on ambiguous questions.

The escalation analysis uses the larger model's cached answers on the same questions. It shows what a router would have achieved with costs counted in weight bytes; a production system with latency, batching and memory limits could see different results.

## 14 Conclusion

Disagreement between two precisions of the same weights is a usable uncertainty signal on these small models, and a modest one. For Qwen3-0.6B it detects wrong answers about as well as the entropy of the answer (0.684 against 0.682), correlates strongly with it, and adds +0.009 [+0.002, +0.015] AUROC when combined with it, mostly on ARC. On the 1.7B model the same score is -0.001 [-0.009, +0.006] from entropy alone and +0.003 [-0.001, +0.006] in combination on the test half. With all questions and cross-fitting the combination gains a small +0.004 [+0.001, +0.006] over the confidence features, concentrated on the HellaSwag subset. The matched-noise control, run on the 0.6B model only, finds no evidence that quantization is special. Random weight noise of the same energy works about as well, and the seed-to-seed spread is large. The best range of copies is the one closest to the original among those run, where the quantizer has done little damage. The signal degrades or reverses for 3-bit copies, and no copy between 6.25 and 8.25 effective bits was tried.

For a local user the practical reading is a ranking by cost. A logistic recalibration of the confidence features is free and calibrates the 0.6B model to 0.014 ECE (entropy alone: 0.032); on the 1.7B model the entropy-only map is not enough (0.073 against 0.021 for the three features). A spare low-precision copy can be added as extra features for a small gain. If one more 8-bit pass is affordable and no larger model is available, re-asking with shifted options is the better use of it. It works as a detector (+0.049 [+0.031, +0.067] AUROC over entropy), and averaging the two answers raises accuracy. If a larger model is available, the scores route differently at equal share (option order best). Once the cost of the signal pass is counted, routing on the free entropy ranking is ahead of both extra-pass scores.

The results come from two small models of one family, plain round-to-nearest quantization, four multiple-choice benchmarks and a simulated cost model, and they may look different with GPTQ or AWQ quantizers, larger models or open-ended answers. The code, cached logits and builder are provided so that another copy, bit-width or benchmark can be added without re-running the rest.

## Appendix A The simulator and the scores, as code

The quantizer, the matched-noise control and the scores are short enough to list in full. The simulation is in memory: nothing is packed, so no speed or memory saving is claimed.

```
def fake_quant(w, bits, g):                       # w: (rows, cols) weight, g divides cols
    x = w.float().reshape(-1, g)                  # one row of the matrix is cut into groups of g
    mn, mx = x.amin(1, keepdim=True), x.amax(1, keepdim=True)
    s = ((mx - mn) / (2 ** bits - 1)).clamp_min(1e-9)
    q = torch.clamp(torch.round((x - mn) / s), 0, 2 ** bits - 1)   # integer code
    return (q * s + mn).reshape(w.shape)          # dequantised weight, stored back as bf16

def matched_noise(w, bits, g=128, per_group=False):  # control: same error energy, no dependence on w
    err = w.float() - fake_quant(w, bits, g)
    sig = err.reshape(-1, g).pow(2).mean(1, keepdim=True).sqrt().expand(-1, g).reshape(w.shape) if per_group
else err.pow(2).mean().sqrt()
    return w.float() + torch.randn_like(w.float()) * sig

def option_probs(letter_logits, n_options):       # softmax over the letters that exist
    z = letter_logits[:n_options]; e = np.exp(z - z.max()); return e / e.sum()

def disagreement(p_hi, p_lo):                     # larger = more likely wrong
    y = p_hi.argmax()
    flip = float(p_lo.argmax() != y); pdrop = 1 - p_lo[y]
    m = (p_hi + p_lo) / 2; H = lambda p: -(p * np.log(np.clip(p, 1e-12, 1))).sum()
    jsd = H(m) - (H(p_hi) + H(p_lo)) / 2
    return flip, pdrop, jsd
```

## Appendix B All configurations

Table 12 lists every copy. The selected copy for each model was chosen on the development half, and the table uses all questions, so its numbers are descriptive and not an out-of-sample estimate.

**Table 12.** Every low-precision copy that was run: accuracy, flips against the 8-bit copy and AUROC on all questions of each disagreement score (1 - p_low[answer], JSD, flip, margin drop, KL).

| model | copy | eff. bits | acc. | flips | 1 - p_low[ans] | JSD | flip | margin drop | KL |
|---|---|---|---|---|---|---|---|---|---|
| Qwen3-0.6B | w3g128 | 3.25 | 24.6% | 74.3% | 0.472 | 0.410 | 0.490 | 0.366 | 0.435 |
| Qwen3-0.6B | w3g64 | 3.50 | 24.3% | 71.9% | 0.448 | 0.386 | 0.472 | 0.355 | 0.400 |
| Qwen3-0.6B | w3g32 | 4.00 | 25.2% | 69.1% | 0.467 | 0.406 | 0.470 | 0.376 | 0.428 |
| Qwen3-0.6B | w4g256 | 4.12 | 39.7% | 48.0% | 0.584 | 0.568 | 0.558 | 0.417 | 0.583 |
| Qwen3-0.6B | w4g128 | 4.25 | 36.8% | 50.6% | 0.550 | 0.477 | 0.537 | 0.360 | 0.499 |
| Qwen3-0.6B | w4g64 | 4.50 | 37.3% | 45.1% | 0.552 | 0.517 | 0.521 | 0.394 | 0.533 |
| Qwen3-0.6B | w4g32 | 5.00 | 39.6% | 44.0% | 0.587 | 0.527 | 0.549 | 0.404 | 0.537 |
| Qwen3-0.6B | w5g128 | 5.25 | 47.3% | 32.4% | 0.666 | 0.578 | 0.583 | 0.434 | 0.580 |
| Qwen3-0.6B | w6g128 | 6.25 | 49.8% | 15.8% | 0.688 | 0.584 | 0.553 | 0.489 | 0.583 |
| Qwen3-1.7B | w4g32 | 5.00 | 60.8% | 27.8% | 0.710 | 0.709 | 0.616 | 0.379 | 0.718 |
| Qwen3-1.7B | w5g128 | 5.25 | 65.1% | 18.5% | 0.746 | 0.724 | 0.599 | 0.366 | 0.729 |
| Qwen3-1.7B | w6g128 | 6.25 | 67.6% | 9.8% | 0.767 | 0.755 | 0.572 | 0.531 | 0.751 |

## Appendix C Protocol details

The prompt has a header line (“The following is a multiple choice question. Choose the correct answer.” or, for HellaSwag, “Complete the sentence by choosing the most plausible ending.”), the question or context, the lettered options one per line, and `Answer:`. The score of option k is the logit of the single token for a space followed by the letter. Prompts longer than 1,024 tokens keep their last 1,024 tokens.

The logistic models standardise features on the training part. The confidence features are the entropy, the negative margin and one minus the maximum probability of the 8-bit copy; the disagreement features are 1 - p_low[answer], JSD, flip and margin drop of the selected copy (and the same four for the shifted-options pass). The fit is by Newton steps with a ridge of 0.001. Cross-fitting uses five folds for pooled and per-task analyses, with each fold held out whole, and two folds for the diagonal of the shift matrices.

The option-order pass is a cyclic shift by one position at 8-bit precision, and the logits are mapped back to the original options before any score is computed. The noise copies have the error energy of the 4-bit group-128 quantizer, per tensor (seeds 0 to 2) or per group (seeds 0 and 1), as listed in Table 3; no 3-bit noise copies were run. For selection, the copy and the score are chosen by AUROC on the development half among 9 copies and five scores for the 0.6B model and among 3 copies for the 1.7B model. Self-consistency was not completed and is not reported.

## Appendix D Reproducibility

The experiment runs in two stages. The first stage runs each model once per configuration and stores, for every question, the logits of the option letters A to E and the full-vocabulary log-normaliser (a 6-number row per question, original option order), and it is resumable per configuration. The configurations are the 16-bit original, the 8-bit group-128 copy, and 3, 4, 5 and 6 bits at group sizes 32, 64, 128 and 256 where listed in the sweep tables (no 7-bit copy was run). The rest are the 8-bit copy with options shifted by one, two and three positions, and the matched-noise copies (per-group and per-tensor variance, 4-bit error energy, seeds 0 to 2 for per-tensor and 0 and 1 for per-group, 0.6B model only). The second stage is this document's builder: every number, table and figure is recomputed from those arrays when the paper is built. The item list is fixed by a seed (ARC full test splits; MMLU 50 per subject with seed 14; HellaSwag 2,000 items with seed 14). The development/test split uses seed 14, cross-fitting folds use seed 0, and bootstrap seeds are fixed.

Software: Python with PyTorch 2.11 (CUDA 12.8), transformers 5.20, NumPy and Matplotlib. The quantizer, the signals, AUROC, risk-coverage, ECE, the logistic fit (Newton steps with a small ridge) and the bootstrap are written from scratch in about 400 lines, so there is no hidden library behaviour. A single RTX 4090 shared with other jobs was used; each configuration of the 0.6B model took between 18 s and 6 min depending on contention, which is why wall-clock times are not reported. The model weights are public; the logits, code and the builder are provided with the paper.